\documentclass[conference]{IEEEtran}
\usepackage{times}

\usepackage[numbers]{natbib}
\usepackage{multicol}
\usepackage[bookmarks=true]{hyperref}

\usepackage{xspace}
\usepackage{xcolor}
\usepackage{bm}
\usepackage{etex}
\usepackage{algpseudocode}
\usepackage{algorithm}

\usepackage{subcaption}

\usepackage{colortbl}
\usepackage{booktabs}
\usepackage{amssymb}
\usepackage{amsmath}
\usepackage{multirow}
\usepackage{gensymb}
\usepackage{float}
\usepackage{graphicx}
\usepackage{xr}
\usepackage[capitalise]{cleveref}

\PassOptionsToPackage{table}{xcolor}

\newcommand{\bdmath}{\begin{dmath}}
\newcommand{\edmath}{\end{dmath}}
\newcommand{\beq}{\begin{equation}}
\newcommand{\eeq}{\end{equation}}
\newcommand{\bdm}{\begin{displaymath}}
\newcommand{\edm}{\end{displaymath}}
\newcommand{\bea}{\begin{eqnarray}}
\newcommand{\eea}{\end{eqnarray}}
\newcommand{\beal}{\beq \begin{array}{ll}}
\newcommand{\eeal}{\end{array} \eeq}
\newcommand{\beas}{\begin{eqnarray*}}
\newcommand{\eeas}{\end{eqnarray*}}
\newcommand{\ba}{\begin{array}}
\newcommand{\ea}{\end{array}}
\newcommand{\bit}{\begin{itemize}}
\newcommand{\eit}{\end{itemize}}
\newcommand{\ben}{\begin{enumerate}}
\newcommand{\een}{\end{enumerate}}

\newcommand{\myParagraph}[1]{{\bf #1.}\xspace}

\newcommand{\hiddenText}{{\color{gray} hidden text.}}
\newcommand{\hideWithText}[1]{\hiddenText}

\newcommand{\Real}[1]{ { {\mathbb R}^{#1} } }
\newcommand{\reals}{\Real{}}

\newcommand{\SEtwo}[1]{\ensuremath{\mathrm{SE}(2)}\xspace}

\newcommand{\SLfour}{\ensuremath{\mathrm{SL}(4)}\xspace}

\newcommand{\Simthree}{\ensuremath{\mathrm{Sim}(3)}\xspace}
\newcommand{\SEthree}{\ensuremath{\mathrm{SE}(3)}\xspace}

\newcommand{\blue}[1]{{\color{blue}#1}}

\newcommand{\linkToPdf}[1]{\href{#1}{\blue{(pdf)}}}
\newcommand{\linkToPpt}[1]{\href{#1}{\blue{(ppt)}}}
\newcommand{\linkToCode}[1]{\href{#1}{\blue{(code)}}}
\newcommand{\linkToWeb}[1]{\href{#1}{\blue{(web)}}}
\newcommand{\linkToVideo}[1]{\href{#1}{\blue{(video)}}}
\newcommand{\linkToMedia}[1]{\href{#1}{\blue{(media)}}}
\newcommand{\award}[1]{\xspace} %

\newcommand{\eg}{\emph{e.g.,}\xspace}
\newcommand{\ie}{\emph{i.e.,}\xspace}
\newcommand{\wrt}{w.r.t.\xspace} 
\newcommand{\name}{VGGT-SLAM 2.0\xspace}

\definecolor{myemerald}{rgb}{0.753, 0.898, 0.804}
\definecolor{mylightgreen}{rgb}{0.894, 0.933, 0.745}
\definecolor{myyellow}{rgb}{0.996, 0.972, 0.780}
\newcommand{\firstc}{\cellcolor{myemerald!100}}
\newcommand{\secondc}{\cellcolor{mylightgreen!100}}

\newcommand{\images}{\mathcal{I}}
\newcommand{\Ks}{\mathcal{K}}
\newcommand{\poses}{\mathcal{T}}

\newcommand{\depths}{\mathcal{D}}
\newcommand{\confs}{\mathcal{C}}
\newcommand{\submap}{\mathcal{S}}

\title{VGGT-SLAM 2.0: Real-time Dense Feed-forward Scene Reconstruction}

\newcommand{\dr}{DUSt3R\xspace}
\newcommand{\mr}{MASt3R\xspace}
\newcommand{\vgs}{VGGT-SLAM\xspace}

\author{Dominic Maggio$^{1}$ and Luca Carlone$^{1*}$}

\begin{document}

\twocolumn[
{   
\renewcommand\twocolumn[1][]{#1}%
\maketitle   
\begin{center}   
    \centering   
    \includegraphics[width=\textwidth]{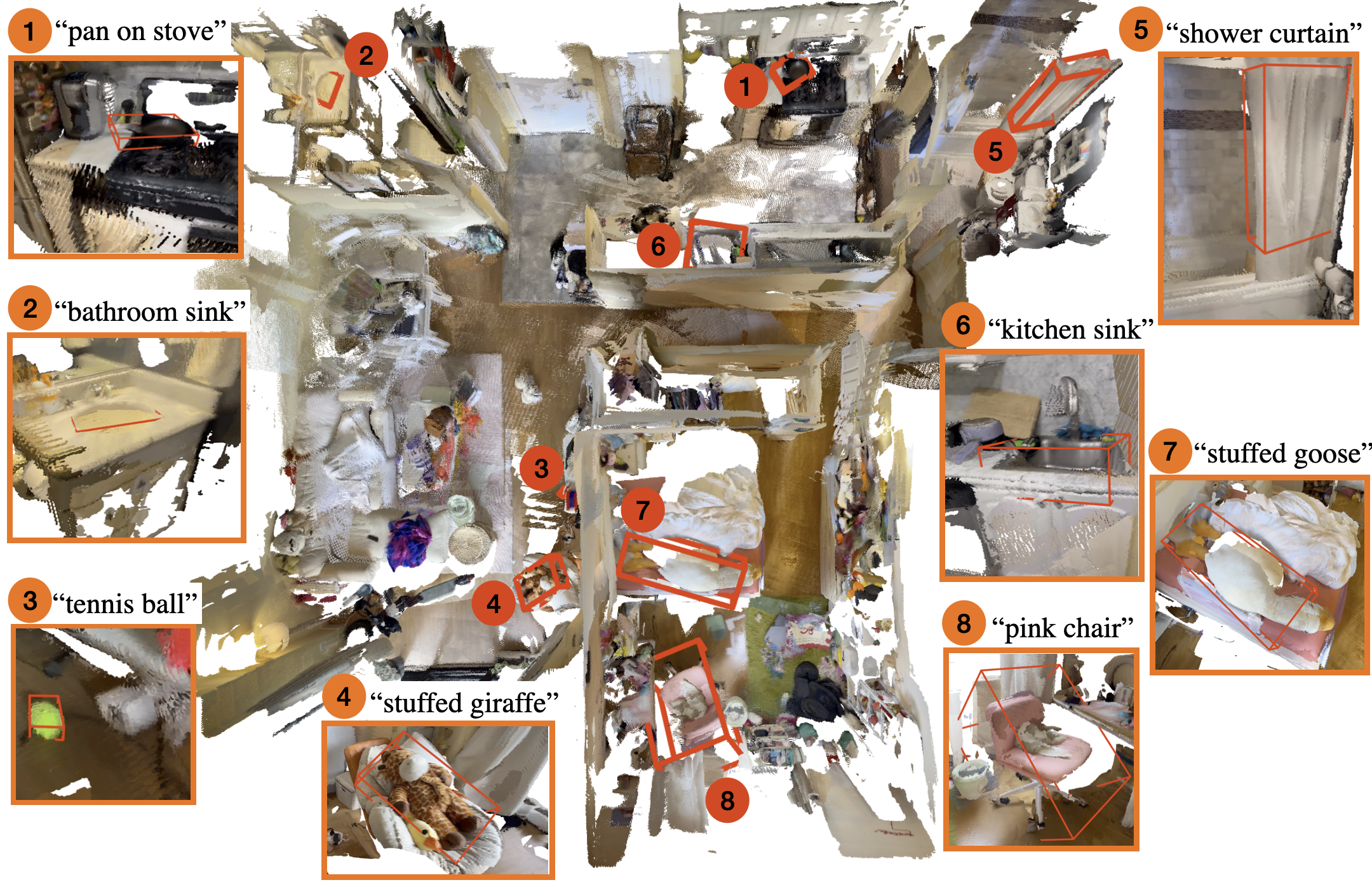}\vspace{-2mm}   
    \captionof{figure}{A \name map of an apartment scene including a living room, kitchen, bedroom, 2 bathrooms, and closet built using 
    uncalibrated RGB images collected with an iPhone. After the map is constructed, it is easy to use it for retrieval of 
    open-set objects. Here, eight example 
    open-set object queries are included, showing the resulting 3D bounding box produced by \name given the provided text query.}   
    \label{fig:cover_fig}   
\end{center}   
}]

\footnotetext[1]{Laboratory for Information \& Decision Systems, Massachusetts Institute of Technology, Cambridge, MA, USA. {\tt\small \{drmaggio, lcarlone\}@mit.edu}}
{\renewcommand{\thefootnote}{*}\footnotetext{Luca holds concurrent appointments as a faculty at the Massachusetts Institute of Technology and as an Amazon Scholar. This paper describes work performed at MIT and is not associated with Amazon. \\
    This work was supported in part by the NSF Graduate Research Fellowship Program under Grant 2141064, the ARL DCIST program, and the ONR RAPID program.}}

\begin{abstract}
We present \name, a real-time RGB feed-forward SLAM system which 
substantially improves upon \vgs for incrementally aligning submaps created from VGGT. 
Firstly, we remove high-dimensional 15-degree-of-freedom drift and planar degeneracy from VGGT-SLAM by 
creating a new factor graph design while still addressing the reconstruction ambiguity of VGGT given unknown 
camera intrinsics. Secondly, by studying the attention layers of VGGT, 
we show that one of the layers is well suited to assist in image retrieval verification for free without additional training, which enables 
both rejecting false positive matches and allows for completing more loop closures. 
Finally, we conduct a suite of experiments which includes showing 
 \name can easily be adapted for open-set 
object detection and demonstrating real-time performance 
while running online onboard a ground robot using a Jetson Thor. We test in environments ranging from cluttered indoor apartments and office scenes 
to a 4,200 square foot barn, and we also demonstrate \name achieves the highest accuracy on the TUM dataset 
with about 23 percent less pose error than \vgs. Code will be released upon publication.
\end{abstract}

\section{Introduction}
\label{sec:intro}

Recently, the foundational task in robotics and computer vision of using images from a camera 
to simultaneously create a 3D reconstruction of a scene and localize the camera has seen a paradigm shift from 
using classical multi-view geometry and optimization techniques~\cite{Carlone25-SLAMHandbook} %
towards building on 
top of feed-forward geometric foundation models~\cite{Wang24cvpr-DUST3R, Leroy24eccv-mast3r, Wang25cvpr-vggt} to develop 
SLAM systems~\cite{Maggio25neurips-VGGT-SLAM, Murai25cvpr-mast3rslam, Zhang25arxiv-vistaslam, Deng25arxiv-vggtlong, Zhou25arxiv-mast3rFusion}. 
These new hybrid SLAM systems, which combine both geometric foundation models with tools from classical SLAM, 
produce a much simpler SLAM system that is both easier to use and maintain. They also produce dense RGB point clouds maps 
while not requiring camera calibration be known beforehand. 

One recent work, \vgs~\cite{Maggio25neurips-VGGT-SLAM}, shows that even though VGGT~\cite{Wang25cvpr-vggt} is unable to 
directly process potentially thousands of frames 
as typically done in SLAM\footnote{VGGT is limited to about 60 frames on an RTX 4090 with 24 GB due to memory constraints.}, it can be extended 
to become a SLAM system by creating and aligning submaps from VGGT where neighboring submaps share an \emph{overlapping frame}. 
The overlapping frame is a common image between submaps which is used for alignment. 
Furthermore, \vgs shows that simply aligning the submaps with 
a similarity transformation (rotation, translation, and scale) is not always sufficient since the VGGT submaps --- which 
rely on uncalibrated camera images --- sometimes have a higher dimensional projective ambiguity. \vgs thus optimizes 
a 15-degree-of-freedom (DoF) alignment on the \SLfour manifold which compensates for submap distortions.

\begin{figure}
    \centering
    \includegraphics[width=.95\linewidth]{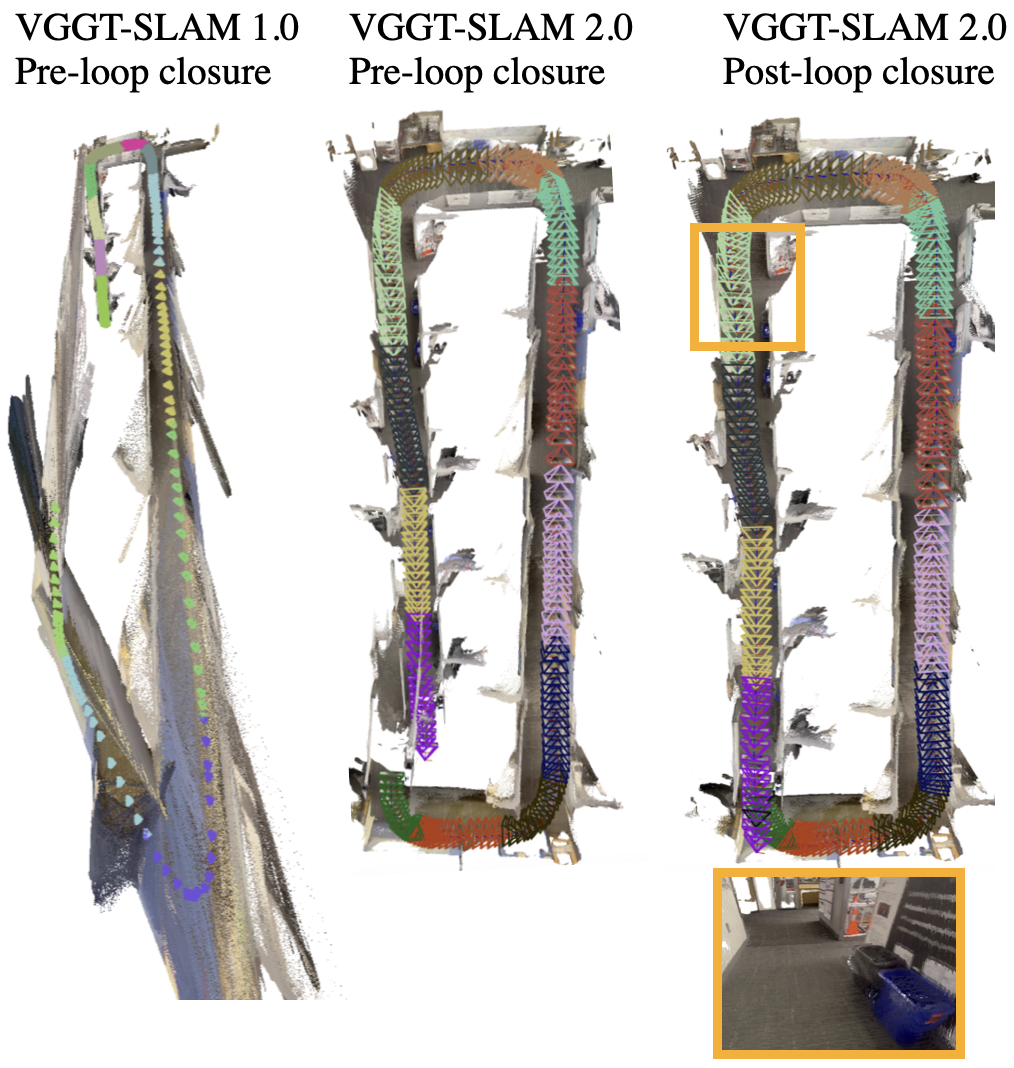}
    \caption{Office loop dataset from~\cite{Maggio25neurips-VGGT-SLAM}. 
    Left: \vgs~\cite{Maggio25neurips-VGGT-SLAM} before detected loop closure showing high-dimensional drift.
    Middle: \name  before detected loop closure showing substantially reduced drift. 
    Right: \name after detected loop closure}
    \label{fig:drift}
    \vspace{-1.5em}
\end{figure}

However, there are several key limitations which hinder the deployment of \vgs which are addressed in this paper. 
The first is the higher dimensional 15-DoF alignment introduces rapid drift between loop closures which can 
severely warp a scene before loop closures (\cref{fig:drift}) and is not always recoverable even with loop closures. 
Solving for the 15-DoF homography is also degenerate in planar scenes which can cause divergence in common 
scenarios such as a camera looking only at a wall or at a flat floor. 

Secondly, the factor graph in \vgs only attempts to estimate the homography between submaps 
instead of correcting keyframe-level errors. This means 
that any potential translation and rotation error from VGGT is handled sub-optimally during global optimization from loop closures. 

Thirdly, image retrieval for loop closure in a VGGT mapping system has largely been treated as an independent task using only information 
from a separate retrieval network. \vgs and VGGT-Long~\cite{Deng25arxiv-vggtlong} for example both use SALAD~\cite{Izquierdo24cvpr-SALAD} to retrieve frames 
while not leveraging information 
inside the VGGT layers to verify the validity of the retrieved frame. 

\myParagraph{Contributions} To address these challenges, we develop \name 
which provides the following contributions:

\begin{itemize}

\item Our first contribution (\Cref{sec:relative}) is to remove the issue of high-dimensional drift and planar degeneracy of VGGT submap 
alignment by directly enforcing that the two overlapping frames in submap alignment must have the same position, rotation, and calibration, and 
solve for a consistent scale factor.

\item Our second contribution (\Cref{sec:relative,sec:loop_closures}) is a new factor graph structure where all keyframes are nodes and the graph contains 
\emph{intra} edges connecting keyframes inside a submap and \emph{inter} edges connecting overlapping frames between submaps. 

\item Our third contribution (\Cref{sec:match}) is to study the attention layers of VGGT to demonstrate that, without any fine-tuning, 
one layer of VGGT is particularly well suited to verify the validity of a potential retrieved frame for loop closures. This helps 
prevent false positives (including in challenging environments such as an office floor with similar instances of desk cubicles), 
and in turn enables establishing more loop closures. 

\item Finally, to show the applicability of \vgs we conduct a suite of experiments showing: 
(1) simple yet effective integration of open-set object detection into the geometric \vgs map, (2) real-time mapping on a Jetson Thor running online 
on board a ground robot, (3) reconstruction of environments ranging from cluttered indoor scenes to a 4,200 
square foot barn to outdoor driving sequences from the Kitti dataset, and (4) the best accuracy of 
recent learning-based methods on the TUM RGB-D dataset with a 23\% error decrease from \vgs. 

\end{itemize}

\section{Related Works}

\myParagraph{Classical mapping techniques}
Classical scene reconstruction methods rely on tracking and associating features~\cite{Zhang15cvpr, Baker04ijcv} across multiple frames 
and then performing backend optimization 
to regress camera poses and scene geometry~\cite{Mouragnon06icra, Schonberger16cvpr-SfMRevisited, Pan24eccv-GLOMAP}. This has been used 
in multiple sparse~\cite{Davison07pami, Qin18tro-vinsmono} and dense~\cite{Engel18pami-DSO, Newcombe2011iccv-dtam} real-time SLAM systems. 
Learning-based methods have also been incorporated in SLAM systems such as for optical flow~\cite{Teed21nips-DROID-SLAM} and 
for neural scene representations~\cite{Zhu22cvpr-niceslam}.
Several works also perform classical optimization of projective 
alignment on the SL(3) group~\cite{Lovegrove12icl-parametric}, 
and~\cite{Madhavan24eccv-projectiveSync} discusses synchronization on the SL(4) group.

\myParagraph{Feed-forward scene reconstruction}
Recently, following the seminal works of \dr~\cite{Wang24cvpr-DUST3R} and \mr~\cite{Leroy24eccv-mast3r}, which are trained to 
take in a pair of uncalibrated monocular images and produce a 3D scene reconstruction, there has been a paradigm shift towards using 
Geometric foundation models (GFM) for SLAM. Since the two aforementioned works can only process a pair of images, multiple works have begun  
developing approaches to process larger numbers of frames. \mr-SFM~\cite{Duisterhof253dv-mast3rSFM} builds on \mr to perform global 
optimization over multiple 
images but quickly grows computationally expensive as the number of frames increases. \mr-SLAM~\cite{Murai25cvpr-mast3rslam} 
is the first work to demonstrate real-time 
SLAM with a GFM. \mr-Fusion~\cite{Zhou25arxiv-mast3rFusion} incorporates IMU and GNSS as 
additional sensor measurements through classical optimization. 

Several works have developed GFMs to directly process multiple 
images~\cite{Wang25arxiv-Cut3R, Wang24arxiv-Spann3R, Keetha25arxiv-mapanything, Wang25cvpr-vggt} with some also 
modifying the networks to also 
enable Gaussian Splatting~\cite{Kerbl23Ttog-GaussianSplatting} reconstructions~\cite{Chen24arxiv-pref3r}. 
However, all of these works 
are restricted by GPU memory usage to a relatively small number of frames. 
To extend VGGT~\cite{Wang25cvpr-vggt} to a large number of frames, VGGT-SLAM~\cite{Maggio25neurips-VGGT-SLAM}
is the first work to incrementally create and align submaps with VGGT. 
VGGT-Long~\cite{Deng25arxiv-vggtlong} aligns VGGT submaps for large-scale 
outdoor self-driving scenes. Recently, SING3R-SLAM~\cite{Li25arxiv-sing3rSLAM} creates submaps and fuses 
them into a global Gaussian Splatting map 
and ViSTA-SLAM~\cite{Zhang25arxiv-vistaslam} uses a lightweight model for two-view association combined with a $\Simthree$ pose graph. 
Additionally, MegaSaM~\cite{Li25cvpr-megasam} uses deep visual SLAM to handle reconstruction for dynamic scenes and TTT3R~\cite{Chen25arxiv-ttt3r} uses Test-Time Training 
to extend CUT3R~\cite{Wang25arxiv-Cut3R} to longer sequences. Concurrently, TALO~\cite{Zhang25arxiv-talo} uses a sparse set of control points and 
Thin Plate Spline deformations fields to align submaps created by a GFM.

\myParagraph{Attention layer analysis}
Perception Encoder~\cite{Bolya25neurips-perceptionEncoder} conducts a study of the layers of a vision encoder 
for downstream tasks. The layers of 
\dr have recently been studied in~\cite{Stary25arxiv-understandingDust3r}, and~\cite{Chen25arxiv-easi3r} shows how an understanding 
of \dr's layers can be used to enable dynamic object detection.
Concurrently, recent 
works have begun exploring the layers of VGGT for 
tasks such as noise suppression~\cite{Han25arxiv-VGGToutliers} and geometric interpretation of 
the layers~\cite{Bratulic25arxiv-VGGTgeometric},
but so far no works have studied the layers of VGGT for image retrieval verification.  %
\section{Notation and Preliminaries}

\myParagraph{Notation}
We represent matrices and vectors with bold letters (\eg \textbf{C}) and sets using 
capital calligraphic fonts (\eg $\confs$). 
To create a submap, $\submap_i$, we pass a set of frames $\images$ to VGGT, and use the output camera calibrations, $\Ks$;
poses, $\poses$; depth maps, $\depths$; and corresponding depth confidence maps, $\confs$. 
All submaps have $n$ keyframes, 
except for the loop closure submaps (\Cref{sec:loop_closures}) which have 2 frames. 

\myParagraph{Preliminaries}
VGGT estimates 3D points which are defined \wrt the first camera of a submap. 3D points can be computed 
from either the point maps 
or by transforming the estimated depth by estimates of the calibrations and poses. 
In this paper, we will instead define 3D points \wrt their respective camera frame. Thus, the points $\textbf{X}$ for frame $i$ (\ie $\textbf{X}_i$) 
are computed through back projection using only $\textbf{K}^{-1}_i$ and $\textbf{D}_i$, 
where $\textbf{X}_i \in~\reals^{3 \times h \times w}$ for image of size $h \times w$.

The $4 \times 4$ homography matrix which expresses the relationship between 
corresponding points $\textbf{X}_i^a,~\textbf{X}_j^b~\in~\reals^{3}$ is: 

\begin{equation}
    \label{eq:homography}
    \textbf{X}_i^a = \textbf{H}^i_j \textbf{X}_{j}^b, 
\end{equation}
where we use overloaded notation such that $\textbf{X}$ is in homogenous coordinates when multiplied by a homography. 
The full $4 \times 4$ homography matrix, 
\begin{equation}
    \label{eq:homography_mat}
    \textbf{H}_i =
    \begin{bmatrix}
    {\textbf{KR}} & \textbf{t} \\
    \mathbf{v}^T & s
    \end{bmatrix},
\end{equation}
has 15 degrees of freedom: 3 for translation (\textbf{t}), 3 for rotation (\textbf{R}), 1 for scale (s), 5 for affine (\ie camera 
calibration, \textbf{K}), and 3 for projective components (\textbf{v})~\cite{Hartley04book}. Since the homography matrix is only up to scale, 
optimizing the residuals between homography 
matrices can be done by mapping each homography to the $\SLfour$ manifold which consist of all $4 \times 4$ matrices 
with unit determinant, allowing for factor graph optimization on the $\SLfour$ manifold~\cite{Maggio25neurips-VGGT-SLAM}. 
More common manifolds used in robotics such as $\SEthree$ and $\Simthree$ are subsets of $\SLfour$.
\section{\name}
\label{sec:method}

Our objective is to incrementally align submaps produced by VGGT into a globally consistent map 
that recovers the true geometry of the scene (up to a similarity transform ambiguity as we do not estimate the true scene scale). 
To do this, in \Cref{sec:relative} we create a factor graph where every node corresponds to a keyframe in a 
submap and edges describe the 
relationship between keyframes. Consecutive submaps share an image such that the first image of a new submap 
is the same as the last image of the prior submap. We refer to the two submaps' respective estimate 
of the frame of this image as the \emph{overlapping frames}. 
In \Cref{sec:match} we show how the attention layers of VGGT can be used for image retrieval verification, 
that is, given a queried frame and a retrieved frame which are predicted to have overlap by a third-party image 
retrieval method (in our case SALAD~\cite{Izquierdo24cvpr-SALAD}), we can use the attention layers to 
provide assurance that VGGT identifies overlap in the images. Finally, in \Cref{sec:loop_closures} we show how our 
new factor graph structure and image retrieval verification are combined to enable global optimization with loop closures.

\begin{figure}
    \centering
    \includegraphics[width=.95\linewidth]{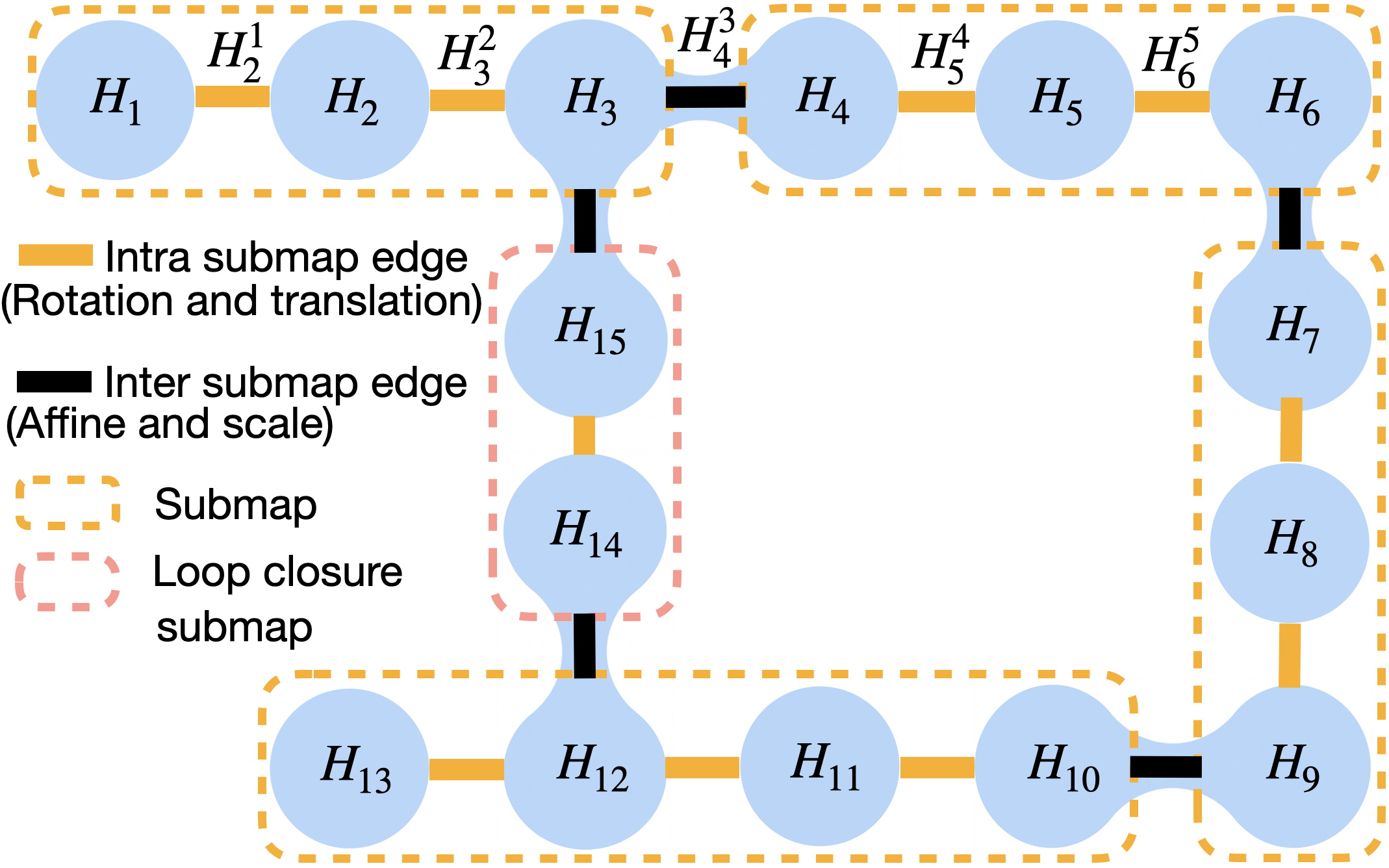}
    \caption{Factor graph structure of \name showing edges inside and between submaps where all keyframes are nodes and loop closure 
    submaps consist of two frames. Nodes connected with an inter submap edge such as 
    $H_3$ and $H_4$ are an example of overlapping nodes (meaning the camera image associated with 
    these nodes is identical).
    Here, for the loop closure submap which is made up of nodes $H_{14}$ and $H_{15}$, 
    $H_{12}$ represents the node of a queried frame which was matched with the node $H_3$ of
    the retrieved frame.}
    \label{fig:pipeline}
    \vspace{-1.5em}
\end{figure}

\subsection{Relative frame alignment}
\label{sec:relative}

For each keyframe, we want to estimate a homography $\textbf{H}_i$ which transforms the 
VGGT reconstruction (which may have submap-specific distortions) 
from the local camera frame to a globally consistent map. 
For edges inside a submap (\textit{intra edges}), 
the edges will only have non-identity rotation and translation (\ie $\SEthree$) variables since 
projective distortion is consistent inside a submap, which means we do not have to apply 
affine correction to each keyframe separately.
Optimizing 
these edges will help correct potential drift in VGGT's translation and rotation estimates.   
For edges between overlapping frames (\textit{inter edges}) there will only be non-identity calibration (\ie affine) 
and scale variables which 
will be used to make each submap's estimate of the scale and calibration of the overlapping frame be identical. 
Since both of these transformations are subsets of $\SLfour$, we use $\SLfour$ nodes for factor graph optimization. 
A visualization of the factor graph structure showing different edge types can be seen in \cref{fig:pipeline}. 
Importantly, given that there is error in VGGT's estimate of calibration, we do not know what the true 
camera calibration is for the overlapping frames --- only that they must have the same calibration. 

\myParagraph{Intra submap frame alignment}
For two keyframes $i$ and $j$ inside a submap, the homography $\mathbf{H}^i_j$ from \cref{eq:homography_mat} that transfers 
correspondences between point clouds $\mathbf{X}_i$ and $\mathbf{X}_j$ 
is simply a rotation and translation which we can get directly from the VGGT outputted poses as:

\begin{equation}
    \label{eq:homography_inner}
    \textbf{H}^i_j = {\textbf{T}_i}^{-1} \textbf{T}_j. 
\end{equation}

\myParagraph{Inter submap frame alignment}
Now, we have to find the alignment between the overlapping frames $i$ and $j$ in overlapping submaps. 
For example, nodes $H_3$ and $H_4$ in \cref{fig:pipeline} correspond to the same keyframe processed in the first and second submap. 
Importantly, instead of solving for a 15-DoF 
transformation by optimizing an alignment over the 3D points corresponding to the two overlapping frames as was done in VGGT-SLAM, 
here we recognize that the overlapping frames by construction must have the same translation, rotation, and camera calibration. Thus, 
we enforce that the relative translation and rotation between the overlapping frames is zero and determine the transformation needed to 
align their calibration. We then solve for the scale factor, $s$. 

The homography in \cref{eq:homography_mat} then simplifies to:
\begin{equation}
    \label{eq:homography_intra}
    \textbf{H}^i_j =
    \begin{bmatrix}
    {\textbf{K}_i}^{-1} \textbf{K}_j & 0 \\
    \mathbf{0}^T & s
    \end{bmatrix},
\end{equation}
where the calibration components, $\textbf{K}_i$ and $\textbf{K}_j$, of the homography come from the VGGT estimated calibrations. Importantly, 
even if the calibrations $\textbf{K}_i$ and $\textbf{K}_j$ do not represent the true camera calibration (which can occur if VGGT incorrectly 
estimates camera calibration for the current submap), this enforces their calibrations to be identical. 
We know they must be identical because the overlapping frames represent the same camera. 
Solving for the scale factor 
is the only time we directly use the VGGT 3D points. To find $s$, 
we leverage two ideas. The first is that since the overlapping frames are identical, corresponding 3D points between them are trivially known. 
The second is that unlike~\cite{Maggio25neurips-VGGT-SLAM}, with our new problem formulation, 
we can easily estimate relative scale separately from other parameters 
of the homography. The 3D points of each 
image are defined not \wrt the first frame of a submap but rather to their own camera frame. 
Therefore, corresponding points seen in overlapping keyframes (\eg $H_3$ and $H_4$ from \cref{fig:pipeline}) must have the same scale. 
To estimate the scale, we warp the point clouds $\mathbf{X}_i$ and $\mathbf{X}_j$ to have the 
same calibration using $\textbf{K}_i^{-1}$ and $\textbf{K}_j$ 
and then estimate the scale factor as the median ratio between corresponding 
3D point distances for all points with sufficient depth confidence $\confs$.

\subsection{Image retrieval verification with VGGT}
\label{sec:match}
\begin{figure}
    \centering
    \begin{subfigure}{\columnwidth}
        \centering
        \includegraphics[width=0.95\columnwidth]{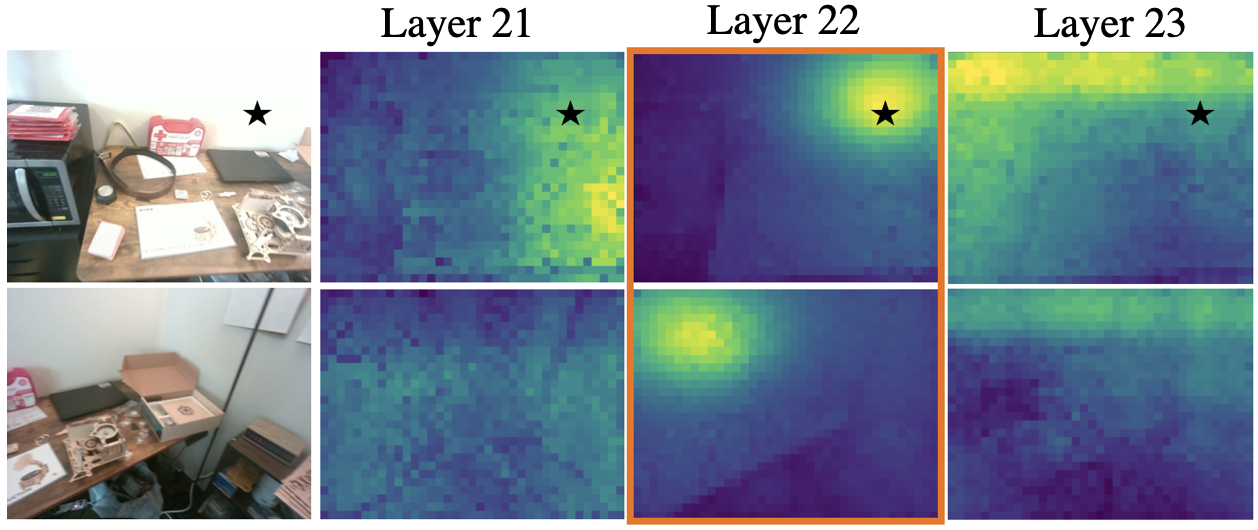}
        \caption{Example images from Clio~\cite{Maggio24ral-clio} apartment scene. Key token placed on a plain wall.}
        \label{fig:layers_a}
    \end{subfigure}
    \vspace{0.5em} %
    \begin{subfigure}{\columnwidth}
        \centering
        \includegraphics[width=0.95\columnwidth]{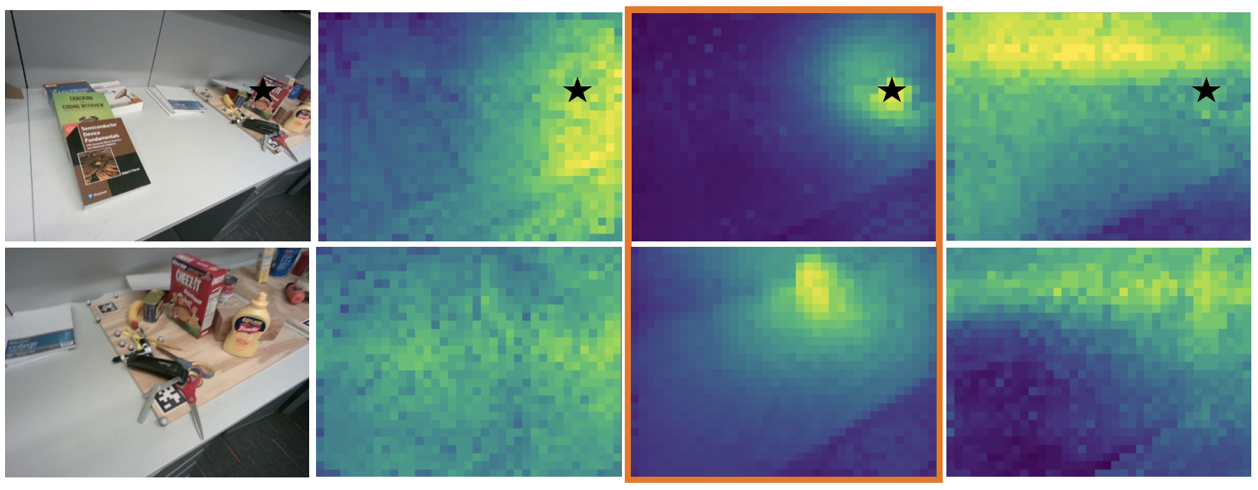}
        \caption{Example images from Clio~\cite{Maggio24ral-clio} cubicle scene. Key token placed on a box.}
        \label{fig:layers_b}
    \end{subfigure}
    \vspace{0.5em} %
    \begin{subfigure}{\columnwidth}
        \centering
        \includegraphics[width=0.95\columnwidth]{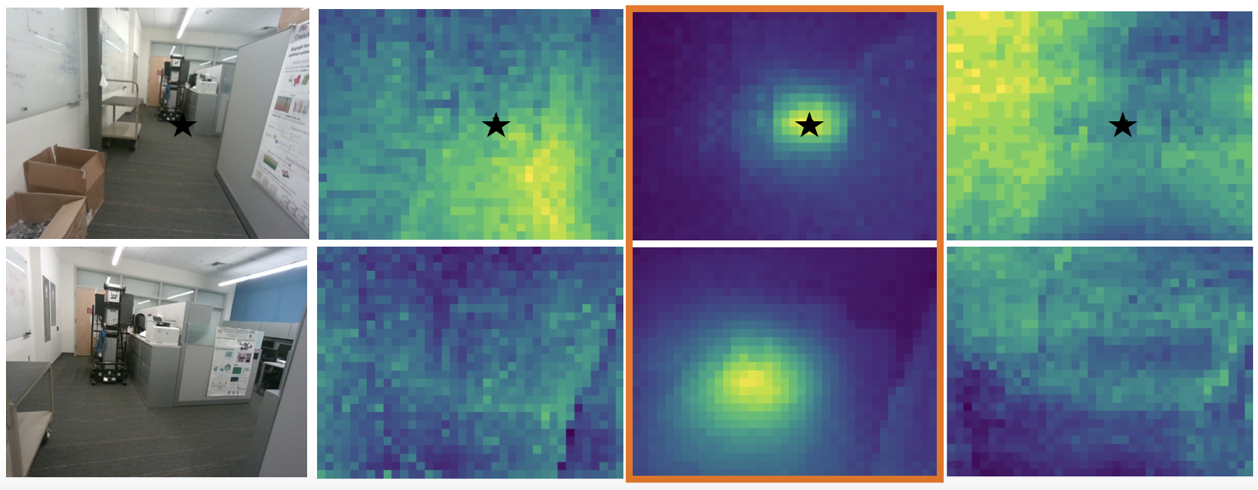}
        \caption{Example images from Clio~\cite{Maggio24ral-clio} office scene. Key token placed on the floor.}
        \label{fig:layers_c}
    \end{subfigure}
    \caption{Attention matrix of query tokens for pairs of images with respect to a selected key token (identified with a black star) using 
    tokens from layers 21, 22, and 23. Layer 22 shows a clear spotlight of attention between corresponding parts of the image pairs.}
    \label{fig:layers}
    \vspace{-1.5em}
\end{figure}

In this section, we demonstrate that by studying the attention layers of VGGT, we can easily extract information to determine 
if VGGT is able to find overlap between input images. While we find our analysis holds for an arbitrary number of frames passed to 
VGGT, since our objective in this section is image 
retrieval verification, we focus on the scenario where VGGT is given two images (a query and a retrieved frame).
In particular, we reveal that layer 22 in VGGT consistently pinpoints whether VGGT predicts correspondences 
between the two images.  In \cref{fig:layers} we visualize the attention matrix of a selected key token (shown with a 
black star), with respect to all other query tokens. Layer 22 consistently shows a spotlight-like attention map which shows larger 
attention values at the correspond location of the image. This is not clearly present in the other layers, including 
the nearby layers (21 and 23) which are shown in \cref{fig:layers} for comparison. This spotlight affects also works on low texture regions 
of a scene as demonstrated by the first example of \cref{fig:layers} which places the query token on a plain white wall.

To leverage this visual observation, we construct a match score in \cref{eq:match_score} between two images 
which captures if 
layer 22 predicts that 
VGGT can reconstruct the two images. 

\begin{equation}
    \label{eq:match_ratios}
    \gamma_t
    = \max\limits_{q \in Q^{(2)}} \left(
    \frac{
    \operatorname{Softmax}\!\left(
    Q^{(2)} {K^{(1)}}^\top
    \right)
    }{
        \max\limits_{q \in Q^{(1)}}
    \operatorname{Softmax}\!\left(
    Q^{(1)} {K^{(1)}}^\top
    \right)
    }
    \right)
\end{equation}

\begin{equation}
    \label{eq:match_score}
    \alpha_{match} = \operatorname{Mean}_{\text{top }25\%} \!\left(\left\{ \gamma_t \right\}\right)
\end{equation}

Here, $Q^{(1)}$ and $K^{(1)}$ represents all query and key tokens (averaged across all heads) for image 1, and $Q^{(2)}$ is likewise 
all query tokens for image 2. 
Intuitively, \cref{eq:match_ratios} computes the attention scores of all query tokens 
of image 2 \wrt the key tokens of image 1 and normalizes by the corresponding attention scores of the query tokens of image 1 \wrt 
the key tokens of image 1, where we use the query tokens that yield the highest attention for each key token.  
Since even a
correctly retrieved pair of images 
will usually only have partial overlap, \cref{eq:match_score} computes the final match score ($\alpha_{match}$) by taking the mean of the 
top 25\% of all ratios.

\begin{figure}[hbt]
    \centering
    \begin{subfigure}{\columnwidth}
        \centering
        \includegraphics[width=0.95\columnwidth]{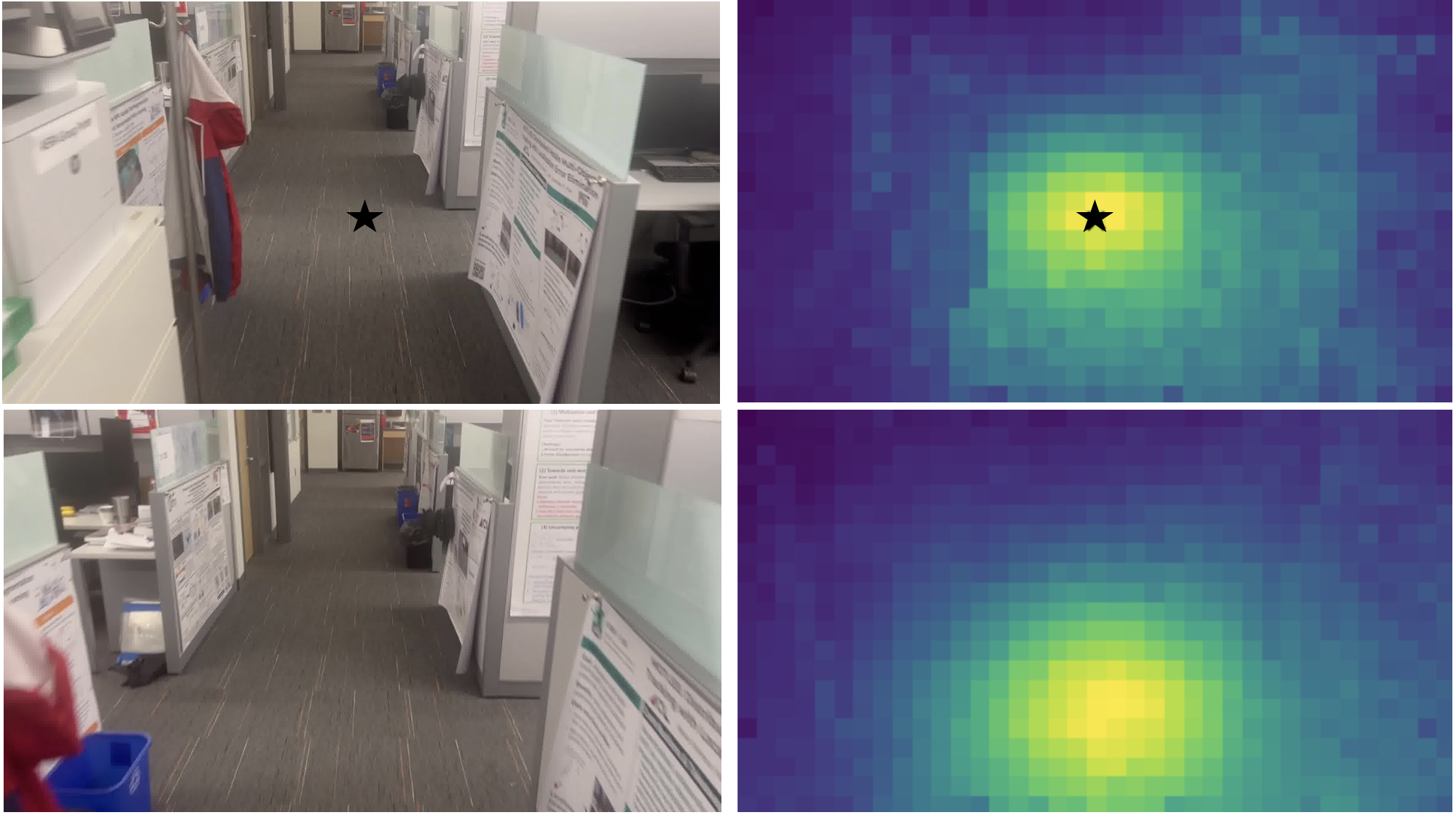}
        \caption{Queried and retrieved frames that have overlap. Our estimated match score: 1.026.}
        \label{fig:match}
    \end{subfigure}
    \vspace{0.5em} %
    \begin{subfigure}{\columnwidth}
        \centering
        \includegraphics[width=0.95\columnwidth]{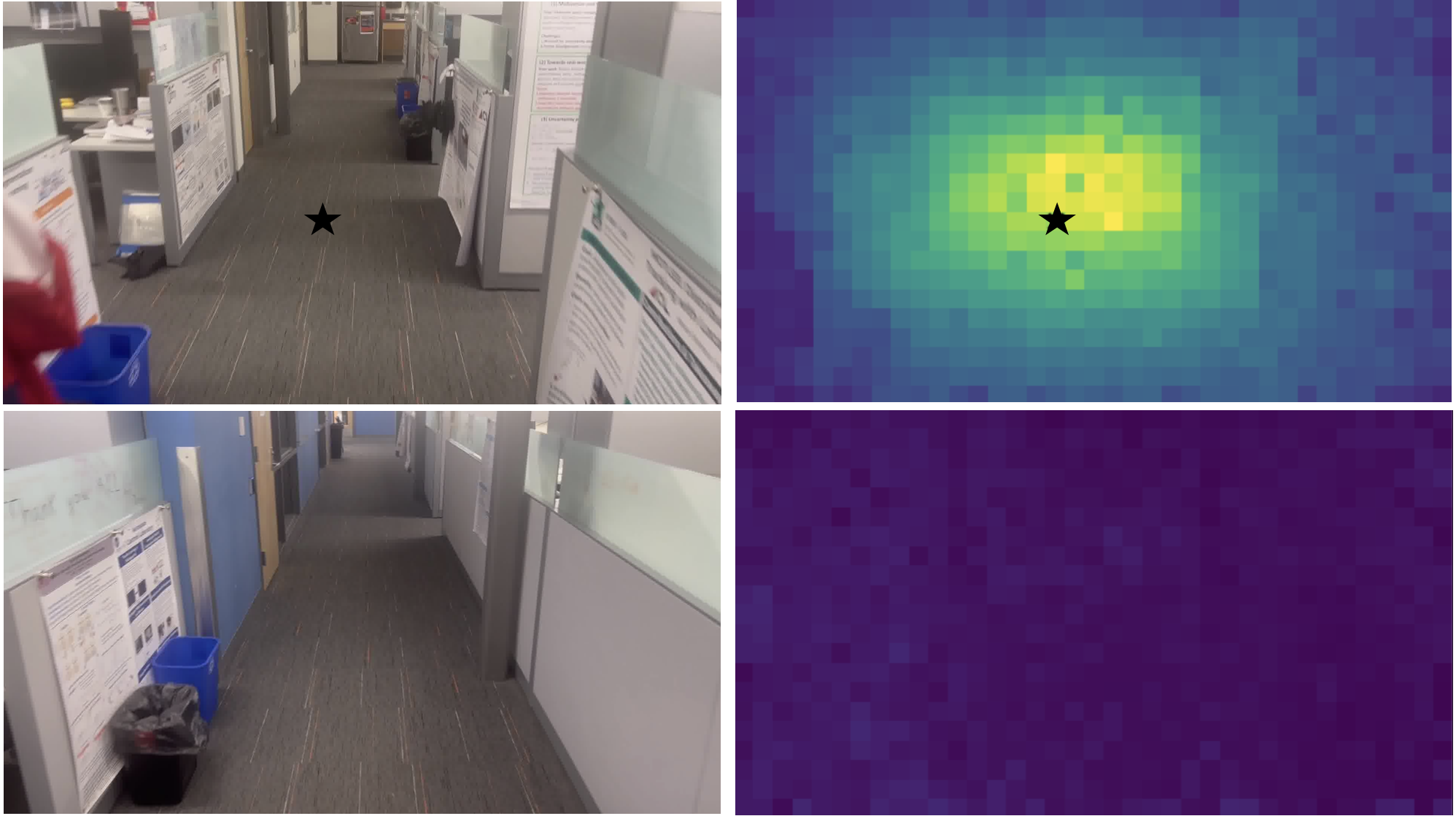}
        \caption{Queried and retrieved frames that do not have overlap. Our estimated match score: 0.491.}
        \label{fig:no_match}
    \end{subfigure}
    \vspace{0.5em} %
    \begin{subfigure}{\columnwidth}
        \centering
        \includegraphics[width=0.95\columnwidth]{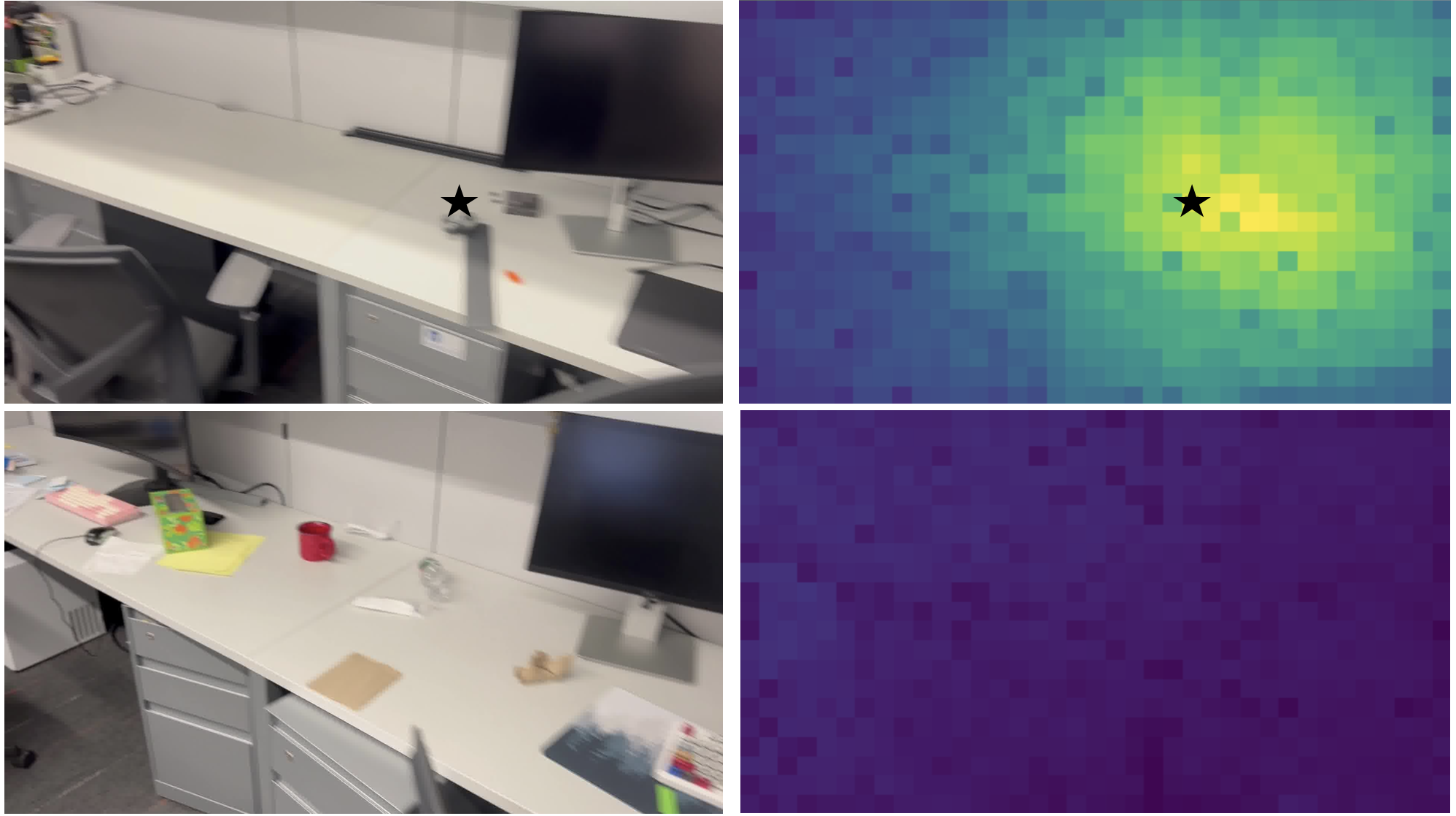}
        \caption{Queried and retrieved frames that do not have overlap, but SALAD incorrectly estimated a match 
        between them. Our estimated match score: 0.550.}
        \label{fig:false_match}
    \end{subfigure}
    \caption{Examples using layer 22 to verify a match between queried and retrieved frames. We visualize attention maps between a 
    selected key token and all query tokens revealing high attention 
    for matching frames and low attention for non-matching frames. This is captured for all pairs of key and query tokens using 
    our match score produced with \cref{eq:match_score}.}
    \label{fig:matches}
    \vspace{-1.5em}
\end{figure}

A visual demonstration of using layer 22 to verify image retrieval is shown in \cref{fig:matches}. Here we show the match score and 
attention map for a sample key token for three cases of query and retried frame: both frames have overlap, both frames do not have overlap, 
and both frames are from similarly looking office cubicles but do not have overlap. 
In this last case, SALAD produced a false positive match with high confidence while our match score accurately shows the frames 
do not overlap. 

\subsection{Loop closures and global optimization}
\label{sec:loop_closures}
Similar to \vgs, we perform image retrieval using SALAD~\cite{Izquierdo24cvpr-SALAD} to check for possible loop closures between frames in the current 
submap and frames in prior submaps (ignoring the most recent prior submap since it will always have overlap with the current submap). 
We differ from \vgs in two key ways. Firstly, instead of adding the retrieved frames into the 
current submap, we create a small submap of two frames (the 
loop closure submap shown in \Cref{fig:pipeline}). This enables greater flexibility as VGGT can be run on the current batch of frames 
before performing image retrieval. This two frame submap consists of the retrieved frame from a prior submap 
and the corresponding query frame in the current submap. This two-frame loop closure submap is passed to VGGT and 
an Inter submap edge is then added between the frame of the loop closure submap and its respective overlapping frame, 
as shown in \Cref{fig:pipeline}.

Secondly, and most importantly, we use information from the attention layers of VGGT to estimate whether the retrieved 
candidate from SALAD is actually a correct match for which VGGT is able to produce a valid estimate as described 
in \cref{sec:match}. This allows us to both relax the 
image retrieval threshold from SALAD to get more potential loop closure candidates and reject false positives in challenging environments. 
Given a SALAD loop closure candidate, we run the procedure in \cref{sec:match} on the queried and retrieved frame and only add a loop 
closure edge if the procedure estimates a match between the two frames.

Given the loop closure constraints, we run global optimization which follows \vgs and is optimized on the \SLfour manifold 
using GTSAM~\cite{gtsam_github} without robust losses. The \SLfour optimization solver has since been added to the official GTSAM release, making installation 
simpler with just a pip install. 

\myParagraph{Recovering the global reconstruction}
Now, given that we have the full set of homographies, we can use them to recover transformed poses and points for our map reconstruction. 

We can recover projection matrices $\mathbf{P_i}$ which transform points from the global frame, $w$, to the camera frame corresponding to image $I_i$ as 

\begin{equation}
    \label{eq:homography_transform}
    \mathbf{P}_i = \mathbf{K}_i^{3 \times 4} (\mathbf{H}^w_i)^{-1}
\end{equation}

Note that differently from VGGT-SLAM, since we have defined points to be relative to their respective frames and not the first 
frame of a submap, the right-hand side of \cref{eq:homography_transform} only has a homography multiplied with a camera calibration (padded with a zeros vector to be $3 \times 4$) 
from VGGT 
instead of additionally being multiplied by a pose from VGGT. Global poses are found by decomposing $\mathbf{P_i}$ into calibration and pose, and global 
3D points are recovered using \cref{eq:homography} with $\mathbf{H}^w_i$.

\section{Experiments}

\subsection{Experimental setup}

To show that \name does not require careful fine-tunning, we keep parameters constant across all quantitative experiments 
and also use the same parameter used by \vgs for minimum disparity (50 pixels) between keyframes and for confidence threshold to 
filter VGGT points using $\confs$ (25\%). Unless otherwise mentioned, all experiments use a SALAD threshold 
of 0.95 (more relaxed than the 0.80 threshold used in~\cite{Maggio25neurips-VGGT-SLAM}) with a retrieval verification threshold of 0.85 
for $\alpha_{match}$.

\subsection{Pose estimation evaluation}
Following the standard pose evaluation protocol which has been used in recent feed-forward SLAM papers 
such as~\cite{Maggio25neurips-VGGT-SLAM, Murai24arxiv-Mast3RSLAM}, we evaluate \name on the TUM RGB-D benchmark 
in \cref{tab:tum}. For fairness, we use submaps of size 32 frames for \name as that is the 
submap size used by \vgs in~\cite{Maggio25neurips-VGGT-SLAM}. \name achieves the best overall average pose error of 4.1 cm which is approximately 23\% lower than \vgs and 
22\% lower than recent ViSTA-SLAM. 

\newcommand{\cg}{\color{gray!70}}

\begin{table*}[t]
    \centering
    \scriptsize
    \begin{tabular}{l|lcccccccccc}
    & &\textbf{360} &\textbf{desk} &\textbf{desk2} &\textbf{floor} &\textbf{plant} &\textbf{room } &\textbf{rpy} &\textbf{teddy} &\textbf{xyz} &\textbf{avg} \\
    \hline
    \multirow{8}{*}{Calibrated} 
    & \cg ORB-SLAM3 \cite{Campos21-TRO} & \cg X & \cg {0.017} & \cg 0.210 & \cg X & \cg 0.034 & \cg X & \cg X & \cg X & \cg \textbf{0.009} & \cg - \\
    & \cg DeepV2D \cite{Teed20iclr-DEEPV2D} & \cg 0.243 & \cg 0.166 & \cg 0.379 & \cg 1.653 & \cg 0.203 & \cg 0.246 & \cg 0.105 & \cg 0.316 & \cg 0.064 & \cg 0.375 \\
    & \cg DeepFactors \cite{Czarnowski20ral-Deepfactors} & \cg 0.159 & \cg 0.170 & \cg 0.253 & \cg 0.169 & \cg 0.305 & \cg 0.364 & \cg 0.043 & \cg 0.601 & \cg 0.035 & \cg 0.233 \\
    & \cg DPV-SLAM \cite{Lipson24eccv-DeepPatch} & \cg 0.112 & \cg 0.018 & \cg 0.029 & \cg 0.057 & \cg 0.021 & \cg 0.330 & \cg 0.030 & \cg 0.084 & \cg {0.010} & \cg 0.076 \\
    & \cg DPV-SLAM++ \cite{Lipson24eccv-DeepPatch} & \cg 0.132 & \cg 0.018 & \cg 0.029 & \cg 0.050 & \cg 0.022 & \cg 0.096 & \cg 0.032 & \cg 0.098 & \cg {0.010} & \cg 0.054 \\
    & \cg GO-SLAM \cite{Zhang23iccv-GOSLAM} & \cg 0.089 & \cg \textbf{0.016} & \cg {0.028} & \cg {0.025} & \cg 0.026 & \cg {0.052} & \cg \textbf{0.019} & \cg 0.048 & \cg {0.010} & \cg {0.035} \\
    & \cg DROID-SLAM \cite{Teed21nips-DROID-SLAM} & \cg 0.111 & \cg 0.018 & \cg 0.042 & \cg \textbf{0.021} & \cg \textbf{0.016} & \cg \textbf{0.049} & \cg {0.026} & \cg 0.048 & \cg 0.012 & \cg 0.038 \\
    & \cg \mr-SLAM \cite{Murai24arxiv-Mast3RSLAM} & \cg \textbf{0.049} & \cg \textbf{0.016} & \cg \textbf{0.024} & \cg {0.025} & \cg {0.020} & \cg 0.061 & \cg 0.027 & \cg 0.041 & \cg \textbf{0.009} & \cg \textbf{0.030} \\
    \hline
    \multirow{5}{*}{Uncalibrated} 
    & DROID-SLAM \cite{Teed21nips-DROID-SLAM} & 0.202 & \secondc 0.032 & 0.091 & \secondc 0.064 & 0.045 & 0.918 & 0.056 & 0.045 & \firstc 0.012 & 0.158 \\
    & \mr-SLAM \cite{Murai24arxiv-Mast3RSLAM} & \secondc 0.070 & 0.035 & 0.055 & \firstc 0.056 & 0.035 & 0.118 & 0.041 & 0.114 & 0.020 & 0.060 \\
    & ViSTA-SLAM~\cite{Zhang25arxiv-vistaslam} & 0.104 & 0.030 & \secondc 0.030 & 0.070 & 0.052 & \secondc 0.067 & \firstc 0.023 & 0.080 & 0.015 & \secondc 0.052 \\
    & \vgs \Simthree~\cite{Maggio25neurips-VGGT-SLAM} & 0.123 & 0.040 & 0.055 & 0.254 & \firstc 0.022 & 0.088 & 0.041 & \firstc \textbf{0.032} & 0.016 & 0.074 \\
    & \vgs \SLfour~\cite{Maggio25neurips-VGGT-SLAM} & 0.071 & \firstc 0.025 & 0.040 & 0.141 & \secondc 0.023 & 0.102 &  0.030 & \secondc 0.034 & \secondc 0.014 & 0.053 \\
    & \name & \firstc 0.050 & \firstc 0.025 & \firstc 0.029 & 0.102 & 0.026 & \firstc 0.063 & \secondc 0.026 & 0.038 & \secondc 0.014 & \firstc 0.041 \\
    \bottomrule
    \end{tabular}
    \caption{Root mean square error~(RMSE) of Absolute trajectory error (ATE (m)) on TUM RGB-D~\cite{Sturm12iros-TUM-RGB-D}.}
    \label{tab:tum}
    \vspace{-1.5em}
\end{table*}

\subsection{Loop closure verification evaluation}

We demonstrate that our image retrieval verification method which 
was presented in \cref{sec:match} can help achieve more loop closures while preventing false positives. In 
\cref{tab:loop_closures} we compare the number of loop closures found by running \name on the three 
scenes of the Clio~\cite{Maggio24ral-clio} datasets 
using a SALAD threshold of 0.80 (which was the value used in~\cite{Maggio25neurips-VGGT-SLAM}) without using retrieval verification 
against the case where we relax the value to 0.95 and use the proposed retrieval verification. The office scene without verification led to false positive loop closures 
causing the reconstruction to diverge even with a SALAD threshold of 0.80 due to challenging similar looking areas such 
as the example from~\cref{fig:false_match}. Meanwhile, using verification led to more loop closures and no false positives, 
even on the office scene. 

To provide an addition experiment on another public dataset, in \cref{tab:retrieval} we compare the Recall@1 score using two retrieval 
methods (SALAD~\cite{Izquierdo24cvpr-SALAD} and NetVLAD~\cite{Cieslewski17mrs-netvlad}) with and without our verification. 
We use the LaMAR~\cite{Sarlin22eccv-lamar} HGE phone validation scene and label a retrieved frame as correct if it is within 25 m of the 
queried frame per standard practice~\cite{Cieslewski17mrs-netvlad}.
In the case 
with verification, we take the top five retrieved frames from SALAD and NetVLAD respectively and score them using our 
$\alpha_{match}$ from \cref{eq:match_score} and return the one with the highest $\alpha_{match}$ to compute the Recall@1 score. We observe 
our verification procedure improves the results of both methods. 

\begin{table}[hbt]
    \centering
    \small
    \setlength{\tabcolsep}{4pt}
    \resizebox{\columnwidth}{!}
    {
    \begin{tabular}{l|cc}
    Dataset  & Without Retrieval Verification & With Retrieval Verification \\
    \midrule
    Clio Cubicle & 2 & \textbf{5}\\
    Clio Apartment & 0 & \textbf{9}\\
    Clio Office & X & \textbf{4}\\
    \bottomrule
    \end{tabular}
    }
    \caption{Number of loop closures on the Clio datasets with and without our retrieval verification. X denotes false positive 
    loop closures were present. }
    \label{tab:loop_closures}
\end{table}

\setlength{\tabcolsep}{2pt}
\begin{table}[hbt]
\centering
\scriptsize
\begin{tabular}{l|cc} %
& Without Retrieval Verification & With Retrieval Verification \\
\hline
SALAD~\cite{Izquierdo24cvpr-SALAD} & 88.45 & \textbf{90.13} \\
NetVLAD~\cite{Cieslewski17mrs-netvlad} & 85.92 & \textbf{89.08} \\
\bottomrule
\end{tabular}
\caption{Recall@1 scores on LaMAR HGE phone dataset showing improved recall with our proposed retrieval verification.}\label{tab:retrieval}
\vspace{-1.5em}
\end{table}

\subsection{Open-set semantic task evaluation}
\label{sec:open_set}
In this section, to further show the usability of \name for robotics, we demonstrate that \name can easily be modified to 
enable 3D open-set object detection.

\begin{figure}[hbt]
    \centering
    \includegraphics[width=.95\linewidth]{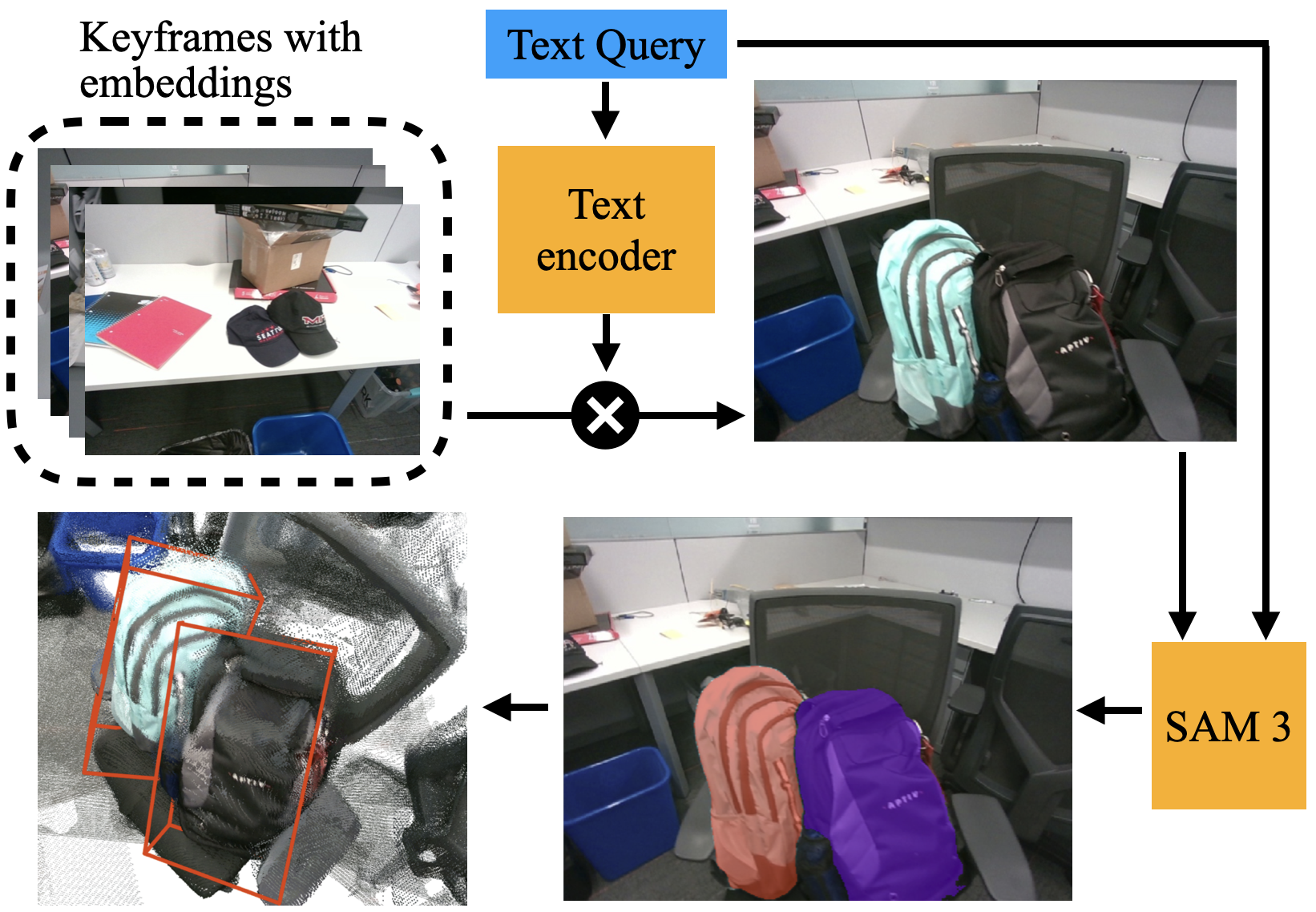}
    \caption{Open-set object detection pipeline where an input text query, in this example ``backpack'', is used to 
    retrieve the best keyframe (which is passed to SAM 3 to perform 2D object segmentation) and finally produce a 3D detection 
    and bounding box of the queried object.}
    \label{fig:semantics_pipeline}
\end{figure}

\myParagraph{Setup}
During mapping, for every keyframe in a submap we compute an image embedding vector using 
the Perception Encoder~\cite{Bolya25neurips-perceptionEncoder} CLIP model. 
To locate a 3D object, the user inputs a text query, from which we compute a CLIP text embedding and locate the corresponding keyframe 
with the highest cosine similarity. 
As in~\cite{Maggio25neurips-VGGT-SLAM}, we have already stored all keyframes to disk memory so 
that they can be used during potential loop closures. We retrieve the corresponding keyframe from memory and pass the 
frame to SAM 3~\cite{Carion25arxiv-sam3} along 
with the text query to perform 2D segmentation. Finally, if SAM 3 detects an object, we identify the points in the \name map 
which correspond to the masked region and compute a minimum 3D oriented bounding box around the points. This process is 
summarized in \cref{fig:semantics_pipeline}. 
The total query time from user text 
input to output 3D bounding box is approximately 0.36 seconds on a 3090 GPU. 

\begin{figure}[hbt]
    \centering
    \includegraphics[width=.95\linewidth]{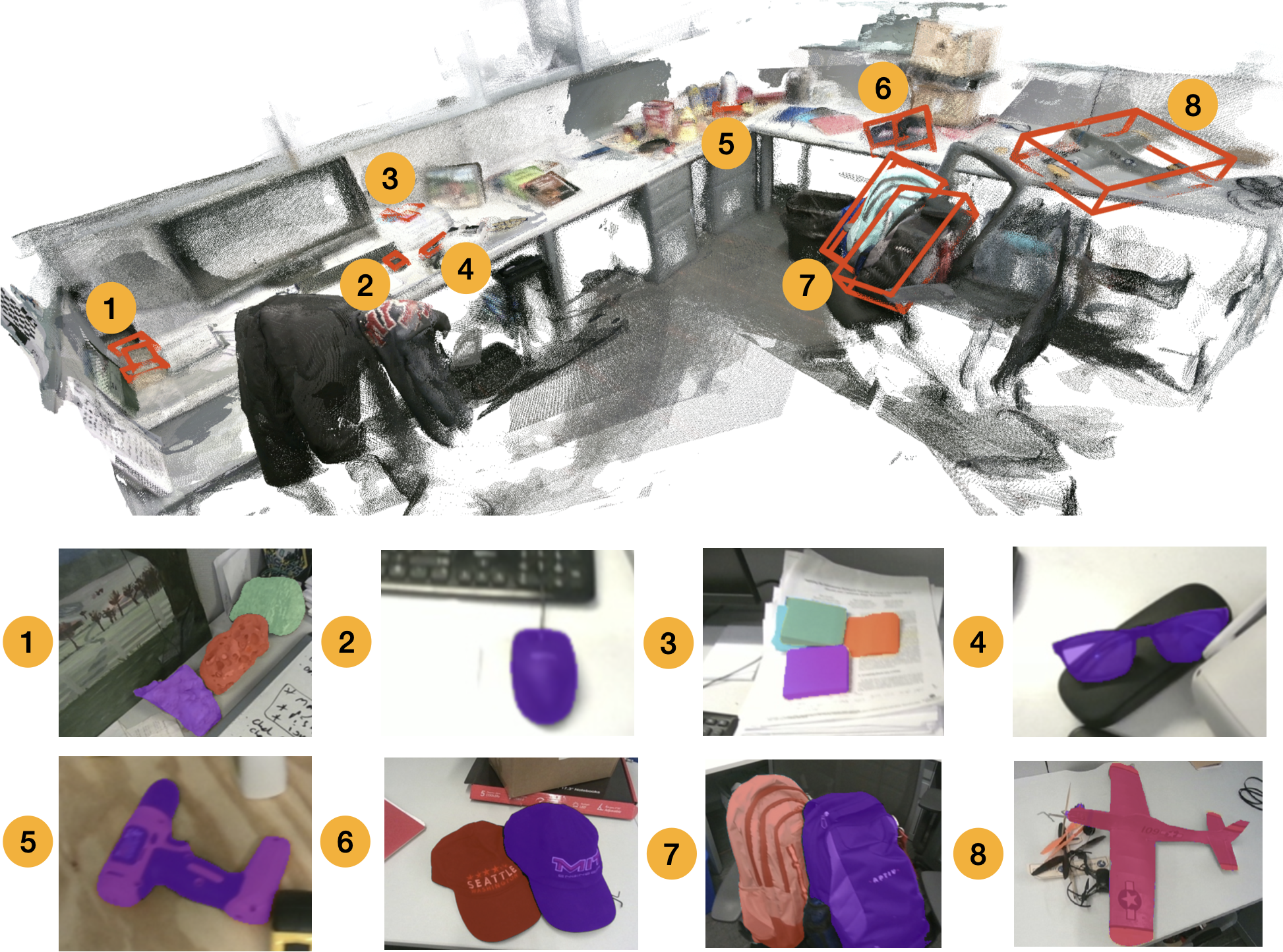}
    \caption{Example 3D open-set object detection results on the Clio cubicle dataset 
    showing estimated 3D oriented bounding boxes and 
    SAM 3 segmentation masks on the queried keyframe. 
    Text input for the 8 queries are: 
    1. rock, 2. mouse, 3. sticky notes, 4. glasses, 5. drill, 6. hat, 7. backpack, and 8. plane.}
    \label{fig:semantics}
\end{figure}

\myParagraph{Qualitative Results}
We provide visual results of multiple objects detected in the \name map through open-set 
querying on the Clio cubicle dataset in \cref{fig:semantics} which uses a RealSense D455 camera and 
in a custom apartment in \cref{fig:cover_fig} using an iPhone camera. We also show a large-scale barn scene 
in \cref{fig:barn} where a large tractor is queried. These results show impressively crisp and reliable object extraction 
using a simple open-set query method built on top of \name.

\subsection{Additional qualitative results}
\begin{figure}[hbt]
    \centering
    \includegraphics[width=.95\linewidth]{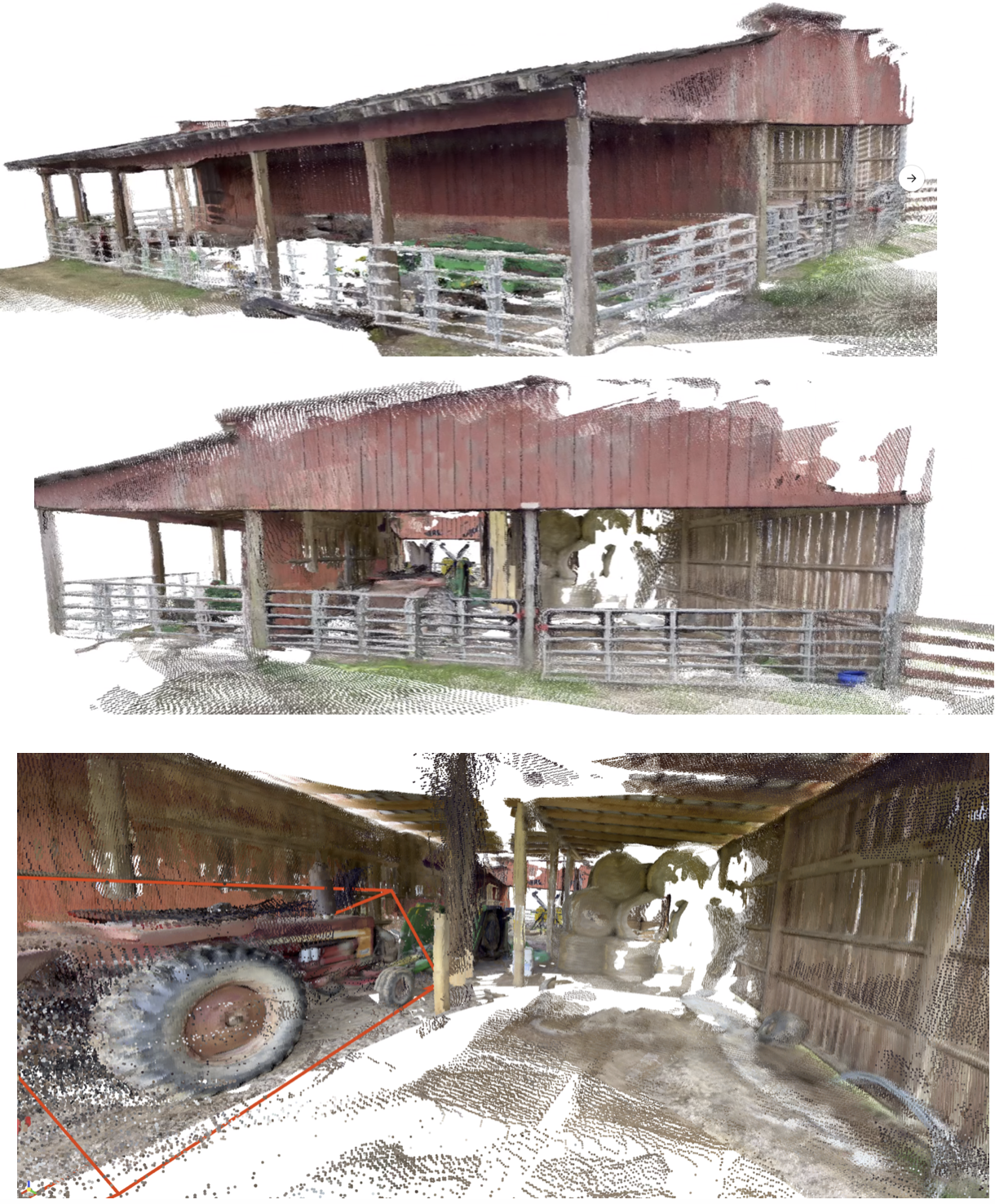}
    \caption{\name reconstruction of a 4200 square foot barn with 34 submaps of size 16 frames using images captured from an iPhone. 
    An example open-set query detecting a tractor 
    inside the barn with the query ``large tractor'' is shown in the bottom figure.}
    \label{fig:barn}
\end{figure}

In this section we provide additional qualitative results to show \name performing in a range of large environments. 
In \cref{fig:barn} we show an example of a reconstruction of the inside and outside a 4200 square foot barn using 
custom data recorded with an iPhone camera. 
We also show an example of querying a large object (a tractor) with our open-set approach 
described in \cref{sec:open_set}.

\begin{figure}[hbt]
    \centering
    \includegraphics[width=.95\linewidth]{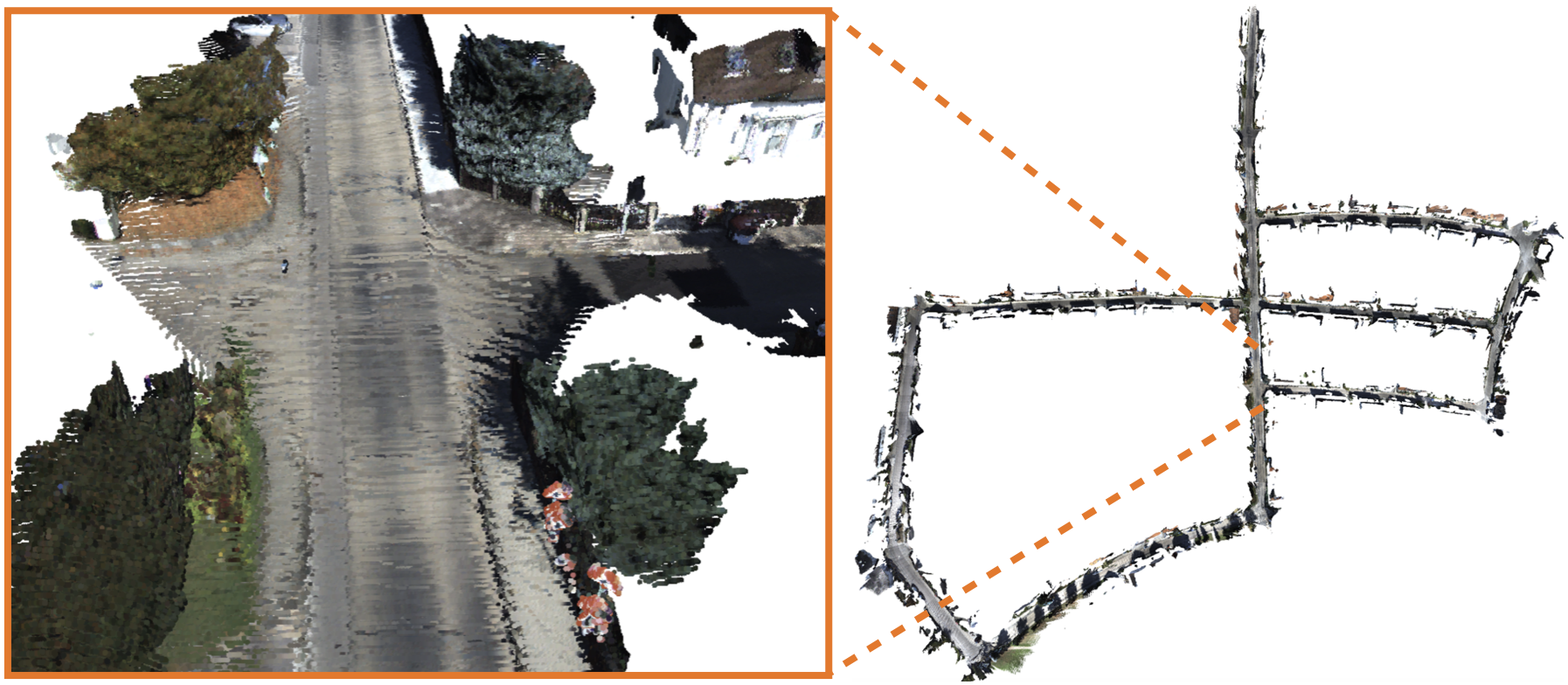}
    \caption{Reconstruction of Kitti outdoor driving sequence 05 which contains 42 submaps of size 32 frames covering 
    a driving distance over 2.2 km.}
    \label{fig:kitti}
    \vspace{-1.5em}
\end{figure}

While we design \name primarily for indoor datasets or small scale outdoor scenes, we also 
include an example from the Kitti~\cite{Geiger12cvpr} 
dataset in \cref{fig:kitti} to provide an example of running on long outdoor sequences to demonstrate improved 
robustness to long sequences. \vgs diverges on the Kitti dataset due to long distances 
between loop closures causing rapid drift, and vulnerability to planar degeneracies as the majority of VGGT reconstructed 
points are on a flat road. 
Both of these scenes (which contain 34 submaps for the barn and 44 for Kitti) are larger than the largest 
scene shown in~\cite{Maggio25neurips-VGGT-SLAM} which 
contained only 22 submaps of size 16 frames. 
Additional visual results are provided in the supplementary video.\footnote{\url{https://www.youtube.com/watch?v=GBdOvd6p4OU}}

\subsection{Real-time experiments and timing results}
The runtime of \name is about 8.4 FPS for a submap of size 16 frames using a 3090 GPU if no open-set CLIP 
vectors are computed, and 6.3 FPS if open-set object detection is enabled. 
FPS is computed by diving the total time to handle a submap by the number of unique frames in the submap (which does not include the overlapping frames). 
A breakdown of the major timing components of \name 
is shown in \cref{tab:timing}. Over half of the time per submap is used to run VGGT (1248 ms) 
which means further reduction in computation can benefit from future faster variants of VGGT. 
Since timing is very hardware specific, we also run \vgs and \mr-SLAM on the same 
machine to provide an approximate reference point for comparison. 
We observe that \mr-SLAM runs at a comparable 7.2 FPS and \vgs runs at 6.9 FPS. 

\begin{table}[hbt]
\centering
\scriptsize
\begin{tabular}{l|cc} %
Stage & total time per submap (ms) \\
\hline
VGGT Inference & 1248 ms \\
Keyframe Detection & 176 ms \\
Loop Closure Detection & 128 ms \\
Backend Optimization & 8 ms \\
Semantics (optional) & 368 ms \\
\bottomrule
\end{tabular}
\caption{Average runtime per submap in milliseconds of the major components of \name. 
Timing results reported using a 3090 GPU for submap size of 16 frames.}\label{tab:timing}
\vspace{-1.5em}
\end{table}

To demonstrate real-time usage of \name for robotics, we 
demonstrate running our entire system onboard a Jetson Thor mounted to a Jackal with a RealSense D455 camera. 
Using submaps of size 4, \name runs at 3.5 FPS onboard the Jetson Thor. 
Our supplementary video 
shows a \name map being created in real-time while the robot explores an office floor. 
Note that by comparison, \vgs is not designed to process a 
live camera stream, only offline images.  %
\section{Limitations}

While our proposed approach substantially improves \vgs and other state-of-the-art methods, it is not without limitations. 
Through experiments on a variety of custom scenes, we did observe some cases of unsatisfactory results and have includes examples in 
the supplementary video to provide both a holistic evaluation and present possible directions of future improvements. 
One such experiment is the 
reconstruction of an empty four-bedroom house which contained keyframes which only saw a plain white wall causing VGGT 
reconstruction to fail. In this case, \name reconstruction will also diverge; possible future work could incorporate 
a lost-tracking module to recover from this. Additionally, our backend factor-graph optimization is only 
over poses and not points. 
While this greatly simplifies the SLAM pipeline (such as by not requiring tracking correspondences) and is 
effective in many of our experiments, in some cases it can cause artifacts from misaligned 3D points and in some challenging scenes leads to  
potentially preventable failure modes. 
Finally, the first frame of the submap is also unequally influential 
since not only does VGGT reconstruct all frames \wrt to first frame, but we also use the first frame to compute the 
scale factor of a submap. Possible future work could optimally select the first frame to both improve scale estimation 
and reduce VGGT failure modes.  %
\section{Conclusion}
We have presented \name, an improved approach to creating and aligning 
submaps with VGGT for dense SLAM with an uncalibrated monocular camera. Our new factor graph design removes the high-dimensional drift 
and planar degeneracies present in \vgs while still accounting for the errors in camera calibration estimated from VGGT. We have conducted 
a study of the attention layers of VGGT and demonstrated how one of the layers can be leveraged for image retrieval verification, enabling 
more loop closures and protection against false positives. Finally, we have conducted a suite of experiments and demonstrations 
such as 
providing an open-set object adaptation of \name and a real time pipeline which can incrementally process a camera stream as a robot 
explores an environment while running onboard a Jetson Thor. Lastly, we remark that since our method does not require any training, 
future improvements to the popular VGGT network such as lighter-weight or more robust variants can easily be plugged in to \name.  
\section*{Acknowledgement} The authors gratefully thank Gary Maggio, Teresa Maggio, and Roubing Liao for assistance with data collection. 

\bibliographystyle{IEEEtran}

\begin{thebibliography}{10}
\providecommand{\url}[1]{#1}
\csname url@samestyle\endcsname
\providecommand{\newblock}{\relax}
\providecommand{\bibinfo}[2]{#2}
\providecommand{\BIBentrySTDinterwordspacing}{\spaceskip=0pt\relax}
\providecommand{\BIBentryALTinterwordstretchfactor}{4}
\providecommand{\BIBentryALTinterwordspacing}{\spaceskip=\fontdimen2\font plus
\BIBentryALTinterwordstretchfactor\fontdimen3\font minus
  \fontdimen4\font\relax}
\providecommand{\BIBforeignlanguage}[2]{{%
\expandafter\ifx\csname l@#1\endcsname\relax
\typeout{** WARNING: IEEEtran.bst: No hyphenation pattern has been}%
\typeout{** loaded for the language `#1'. Using the pattern for}%
\typeout{** the default language instead.}%
\else
\language=\csname l@#1\endcsname
\fi
#2}}
\providecommand{\BIBdecl}{\relax}
\BIBdecl

\bibitem{Carlone25-SLAMHandbook}
L.~Carlone, A.~Kim, T.~Barfoot, D.~Cremers, and F.~Dellaert, Eds., \emph{{SLAM
  Handbook.} From Localization and Mapping to Spatial Intelligence}.\hskip 1em
  plus 0.5em minus 0.4em\relax Cambridge University Press, 2026.

\bibitem{Wang24cvpr-DUST3R}
S.~Wang, V.~Leroy, Y.~Cabon, B.~Chidlovskii, and J.~Revaud, ``Dust3r: Geometric
  3d vision made easy,'' in \emph{IEEE Conf. on Computer Vision and Pattern
  Recognition (CVPR)}, 2024, pp. 20\,697--20\,709.

\bibitem{Leroy24eccv-mast3r}
V.~Leroy, Y.~Cabon, and J.~Revaud, ``Grounding image matching in 3d with
  mast3r,'' in \emph{European Conf. on Computer Vision (ECCV)}, 2024.

\bibitem{Wang25cvpr-vggt}
J.~Wang, M.~Chen, N.~Karaev, A.~Vedaldi, C.~Rupprecht, and D.~Novotny, ``Vggt:
  Visual geometry grounded transformer,'' in \emph{IEEE Conf. on Computer
  Vision and Pattern Recognition (CVPR)}, 2025.

\bibitem{Maggio25neurips-VGGT-SLAM}
D.~Maggio, H.~Lim, and L.~Carlone, ``{VGGT-SLAM}: Dense {RGB SLAM} optimized on
  the {SL(4)} manifold,'' in \emph{Conf. on Neural Information Processing
  Systems (NeurIPS)}, 2025,
  \linkToCode{https://github.com/MIT-SPARK/VGGT-SLAM}.

\bibitem{Murai25cvpr-mast3rslam}
R.~Murai, E.~Dexheimer, and A.~J. Davison, ``Mast3r-slam: Real-time dense slam
  with 3d reconstruction priors,'' in \emph{IEEE Conf. on Computer Vision and
  Pattern Recognition (CVPR)}, 2025, pp. 16\,695--16\,705.

\bibitem{Zhang25arxiv-vistaslam}
G.~Zhang, S.~Qian, X.~Wang, and D.~Cremers, ``Vista-slam: Visual slam with
  symmetric two-view association,'' \emph{arXiv preprint arXiv:2509.01584},
  2025.

\bibitem{Deng25arxiv-vggtlong}
K.~Deng, Z.~Ti, J.~Xu, J.~Yang, and J.~Xie, ``Vggt-long: Chunk it, loop it,
  align it--pushing vggt's limits on kilometer-scale long rgb sequences,''
  \emph{arXiv preprint arXiv:2507.16443}, 2025.

\bibitem{Zhou25arxiv-mast3rFusion}
Y.~Zhou, X.~Li, S.~Li, Z.~Yan, C.~Xia, and S.~Feng, ``Mast3r-fusion:
  Integrating feed-forward visual model with imu, gnss for high-functionality
  slam,'' \emph{arXiv preprint arXiv:2509.20757}, 2025.

\bibitem{Izquierdo24cvpr-SALAD}
S.~Izquierdo and J.~Civera, ``Optimal transport aggregation for visual place
  recognition,'' in \emph{IEEE Conf. on Computer Vision and Pattern Recognition
  (CVPR)}, June 2024.

\bibitem{Zhang15cvpr}
G.~Zhang and P.~Vela, ``Good features to track for visual slam,'' in \emph{IEEE
  Conf. on Computer Vision and Pattern Recognition (CVPR)}, 2015.

\bibitem{Baker04ijcv}
S.~Baker and I.~Matthews, ``Lucas-kanade 20 years on: A unifying framework,''
  \emph{Intl. J. of Computer Vision}, vol.~56, no.~3, pp. 221--255, 2004.

\bibitem{Mouragnon06icra}
E.~Mouragnon, M.~Lhuillier, M.~Dhome, F.~Dekeyser, and P.~Sayd, ``3d
  reconstruction of complex structures with bundle adjustment: an incremental
  approach,'' in \emph{IEEE Intl. Conf. on Robotics and Automation (ICRA)}, May
  2006, pp. 3055--3061.

\bibitem{Schonberger16cvpr-SfMRevisited}
J.~L. Schonberger and J.-M. Frahm, ``Structure-from-motion revisited,'' in
  \emph{IEEE Conf. on Computer Vision and Pattern Recognition (CVPR)}, 2016,
  pp. 4104--4113.

\bibitem{Pan24eccv-GLOMAP}
L.~Pan, D.~Barath, M.~Pollefeys, and J.~L. Sch\"{o}nberger, ``{Global
  Structure-from-Motion Revisited},'' in \emph{European Conf. on Computer
  Vision (ECCV)}, 2024.

\bibitem{Davison07pami}
A.~Davison, I.~Reid, N.~Molton, and O.~Stasse, ``Mono{SLAM}: Real-time single
  camera {SLAM},'' \emph{{IEEE} Trans. Pattern Anal. Machine Intell.}, vol.~29,
  no.~6, pp. 1052--1067, Jun 2007.

\bibitem{Qin18tro-vinsmono}
T.~Qin, P.~Li, and S.~Shen, ``Vins-mono: A robust and versatile monocular
  visual-inertial state estimator,'' \emph{IEEE Transactions on Robotics},
  vol.~34, no.~4, pp. 1004--1020, 2018.

\bibitem{Engel18pami-DSO}
J.~Engel, V.~Koltun, and D.~Cremers, ``Direct sparse odometry,'' \emph{{IEEE}
  Trans. Pattern Anal. Machine Intell.}, 2018.

\bibitem{Newcombe2011iccv-dtam}
R.~A. Newcombe, S.~J. Lovegrove, and A.~J. Davison, ``Dtam: Dense tracking and
  mapping in real-time,'' in \emph{Intl. Conf. on Computer Vision
  (ICCV)}.\hskip 1em plus 0.5em minus 0.4em\relax IEEE, 2011, pp. 2320--2327.

\bibitem{Teed21nips-DROID-SLAM}
Z.~Teed and J.~Deng, ``{DROID}-{SLAM}: Deep visual {SLAM} for monocular,
  stereo, and {RGB}-d cameras,'' in \emph{Advances in Neural Information
  Processing Systems (NIPS)}, A.~Beygelzimer, Y.~Dauphin, P.~Liang, and J.~W.
  Vaughan, Eds., 2021.

\bibitem{Zhu22cvpr-niceslam}
Z.~Zhu, S.~Peng, V.~Larsson, W.~Xu, H.~Bao, Z.~Cui, M.~R. Oswald, and
  M.~Pollefeys, ``{NICE-SLAM}: Neural implicit scalable encoding for slam,'' in
  \emph{IEEE Conf. on Computer Vision and Pattern Recognition (CVPR)}, June
  2022.

\bibitem{Lovegrove12icl-parametric}
S.~Lovegrove, ``Parametric dense visual {SLAM},'' Ph.D. dissertation, 2012.

\bibitem{Madhavan24eccv-projectiveSync}
R.~Madhavan, A.~Fusiello, and F.~Arrigoni, ``Synchronization of projective
  transformations,'' in \emph{European Conf. on Computer Vision (ECCV)},
  A.~Leonardis, E.~Ricci, S.~Roth, O.~Russakovsky, T.~Sattler, and G.~Varol,
  Eds.\hskip 1em plus 0.5em minus 0.4em\relax Springer Nature Switzerland,
  2025, pp. 18--36.

\bibitem{Duisterhof253dv-mast3rSFM}
B.~P. Duisterhof, L.~Zust, P.~Weinzaepfel, V.~Leroy, Y.~Cabon, and J.~Revaud,
  ``Mast3r-sfm: a fully-integrated solution for unconstrained
  structure-from-motion,'' in \emph{2025 International Conference on 3D Vision
  (3DV)}.\hskip 1em plus 0.5em minus 0.4em\relax IEEE, 2025, pp. 1--10.

\bibitem{Wang25arxiv-Cut3R}
Q.~Wang, Y.~Zhang, A.~Holynski, A.~A. Efros, and A.~Kanazawa, ``{Continuous 3D
  Perception Model with Persistent State},'' \emph{arXiv preprint
  arXiv:2501.12387}, 2025.

\bibitem{Wang24arxiv-Spann3R}
H.~Wang and L.~Agapito, ``{3D reconstruction with spatial memory},''
  \emph{arXiv preprint arXiv:2408.16061}, 2024.

\bibitem{Keetha25arxiv-mapanything}
N.~Keetha, N.~M{\"u}ller, J.~Sch{\"o}nberger, L.~Porzi, Y.~Zhang, T.~Fischer,
  A.~Knapitsch, D.~Zauss, E.~Weber, N.~Antunes \emph{et~al.}, ``Mapanything:
  Universal feed-forward metric 3d reconstruction,'' \emph{arXiv preprint
  arXiv:2509.13414}, 2025.

\bibitem{Kerbl23Ttog-GaussianSplatting}
B.~Kerbl, G.~Kopanas, T.~Leimk{\"u}hler, and G.~Drettakis, ``3d gaussian
  splatting for real-time radiance field rendering,'' \emph{ACM Transactions on
  Graphics}, vol.~42, no.~4, July 2023.

\bibitem{Chen24arxiv-pref3r}
Z.~Chen, J.~Yang, and H.~Yang, ``Pref3r: Pose-free feed-forward 3d gaussian
  splatting from variable-length image sequence,'' \emph{arXiv preprint
  arXiv:2411.16877}, 2024.

\bibitem{Li25arxiv-sing3rSLAM}
K.~Li, M.~Niemeyer, S.~Wang, S.~Gasperini, N.~Navab, and F.~Tombari,
  ``Sing3r-slam: Submap-based indoor monocular gaussian slam with 3d
  reconstruction priors,'' \emph{arXiv preprint arXiv:2511.17207}, 2025.

\bibitem{Li25cvpr-megasam}
Z.~Li, R.~Tucker, F.~Cole, Q.~Wang, L.~Jin, V.~Ye, A.~Kanazawa, A.~Holynski,
  and N.~Snavely, ``Megasam: Accurate, fast and robust structure and motion
  from casual dynamic videos,'' in \emph{IEEE Conf. on Computer Vision and
  Pattern Recognition (CVPR)}, 2025, pp. 10\,486--10\,496.

\bibitem{Chen25arxiv-ttt3r}
X.~Chen, Y.~Chen, Y.~Xiu, A.~Geiger, and A.~Chen, ``Ttt3r: 3d reconstruction as
  test-time training,'' \emph{arXiv preprint arXiv:2509.26645}, 2025.

\bibitem{Zhang25arxiv-talo}
F.~Zhang, T.~Zhang, K.~Khosoussi, Z.~Zhang, Z.~Huang, and Y.~Luo, ``Talo:
  Pushing 3d vision foundation models towards globally consistent online
  reconstruction,'' \emph{arXiv preprint arXiv:2512.02341}, 2025.

\bibitem{Bolya25neurips-perceptionEncoder}
D.~Bolya, P.-Y. Huang, P.~Sun, J.~H. Cho, A.~Madotto, C.~Wei, T.~Ma, J.~Zhi,
  J.~Rajasegaran, H.~Rasheed, J.~Wang, M.~Monteiro, H.~Xu, S.~Dong, N.~Ravi,
  D.~Li, P.~Doll{\'a}r, and C.~Feichtenhofer, ``Perception encoder: The best
  visual embeddings are not at the output of the network,'' in \emph{Advances
  in Neural Information Processing Systems (NeurIPS)}, 2025.

\bibitem{Stary25arxiv-understandingDust3r}
M.~Stary, J.~Gaubil, A.~Tewari, and V.~Sitzmann, ``Understanding multi-view
  transformers,'' \emph{arXiv preprint arXiv:2510.24907}, 2025.

\bibitem{Chen25arxiv-easi3r}
X.~Chen, Y.~Chen, Y.~Xiu, A.~Geiger, and A.~Chen, ``Easi3r: Estimating
  disentangled motion from dust3r without training,'' \emph{arXiv preprint
  arXiv:2503.24391}, 2025.

\bibitem{Han25arxiv-VGGToutliers}
J.~Han, S.~Hong, J.~Jung, W.~Jang, H.~An, Q.~Wang, S.~Kim, and C.~Feng,
  ``Emergent outlier view rejection in visual geometry grounded transformers,''
  \emph{arXiv preprint arXiv:2512.04012}, 2025.

\bibitem{Bratulic25arxiv-VGGTgeometric}
J.~Bratuli{\'c}, S.~Mittal, T.~Brox, and C.~Rupprecht, ``On geometric
  understanding and learned data priors in vggt,'' \emph{arXiv preprint
  arXiv:2512.11508}, 2025.

\bibitem{Hartley04book}
R.~I. Hartley and A.~Zisserman, \emph{Multiple View Geometry in Computer
  Vision}, 2nd~ed.\hskip 1em plus 0.5em minus 0.4em\relax Cambridge University
  Press, 2004.

\bibitem{Maggio24ral-clio}
D.~Maggio, Y.~Chang, N.~Hughes, M.~Trang, D.~Griffith, C.~Dougherty,
  E.~Cristofalo, L.~Schmid, and L.~Carlone, ``Clio: Real-time task-driven
  open-set {3D} scene graphs,'' \emph{{IEEE} Robotics and Automation Letters
  ({RA-L})}, vol.~9, no.~10, pp. 8921--8928, 2024,
  \linkToPdf{https://arxiv.org/pdf/2404.13696},\linkToVideo{https://youtu.be/m-HJO10qhSQ},\linkToWeb{https://news.mit.edu/2024/helping-robots-focus-on-objects-that-matter-0930}.

\bibitem{gtsam_github}
\BIBentryALTinterwordspacing
F.~Dellaert and G.~Contributors, ``borglab/gtsam,'' May 2022. [Online].
  Available: \url{https://github.com/borglab/gtsam)}
\BIBentrySTDinterwordspacing

\bibitem{Murai24arxiv-Mast3RSLAM}
R.~Murai, E.~Dexheimer, and A.~J. Davison, ``{MASt3R-SLAM: Real-Time Dense SLAM
  with 3D Reconstruction Priors},'' \emph{arXiv preprint arXiv:2412.12392},
  2024.

\bibitem{Campos21-TRO}
C.~Campos, R.~Elvira, J.~J.~G. Rodr{\'\i}guez, J.~M. Montiel, and J.~D.
  Tard{\'o}s, ``{ORB-SLAM3}: An accurate open-source library for visual,
  visual--inertial, and multimap {SLAM},'' \emph{{IEEE} Trans. Robotics}, 2021.

\bibitem{Teed20iclr-DEEPV2D}
Z.~Teed and J.~Deng, ``{DEEPV2D: Video to depth with differentiable structure
  from motion},'' \emph{Intl. Conf. on Learning Representations (ICLR)}, 2018.

\bibitem{Czarnowski20ral-Deepfactors}
J.~Czarnowski, T.~Laidlow, R.~Clark, and A.~Davison, ``{DeepFactors}: Real-time
  probabilistic dense monocular {SLAM},'' \emph{{IEEE} Robotics and Automation
  Letters}, vol.~5, no.~2, pp. 721--728, 2020.

\bibitem{Lipson24eccv-DeepPatch}
L.~Lipson, Z.~Teed, and J.~Deng, ``{Deep patch visual SLAM},'' in
  \emph{European Conf. on Computer Vision (ECCV)}, 2024, pp. 424--440.

\bibitem{Zhang23iccv-GOSLAM}
Y.~Zhang, F.~Tosi, S.~Mattoccia, and M.~Poggi, ``{GO-SLAM: Global optimization
  for consistent 3D instant reconstruction},'' in \emph{Intl. Conf. on Computer
  Vision (ICCV)}, 2023, pp. 3727--3737.

\bibitem{Sturm12iros-TUM-RGB-D}
J.~Sturm, N.~Engelhard, F.~Endres, W.~Burgard, and D.~Cremers, ``A benchmark
  for the evaluation of {RGB-D} {SLAM} systems,'' in \emph{IEEE/RSJ Intl. Conf.
  on Intelligent Robots and Systems (IROS)}.\hskip 1em plus 0.5em minus
  0.4em\relax IEEE, 2012, pp. 573--580.

\bibitem{Cieslewski17mrs-netvlad}
T.~{Cieslewski} and D.~{Scaramuzza}, ``Efficient decentralized visual place
  recognition from full-image descriptors,'' in \emph{2017 International
  Symposium on Multi-Robot and Multi-Agent Systems (MRS)}, 2017, pp. 78--82.

\bibitem{Sarlin22eccv-lamar}
P.-E. Sarlin, M.~Dusmanu, J.~L. Sch\"onberger, P.~Speciale, L.~Gruber,
  V.~Larsson, O.~Miksik, and M.~Pollefeys, ``{LaMAR: Benchmarking Localization
  and Mapping for Augmented Reality},'' in \emph{European Conf. on Computer
  Vision (ECCV)}, 2022.

\bibitem{Carion25arxiv-sam3}
\BIBentryALTinterwordspacing
N.~Carion, L.~Gustafson, Y.-T. Hu, S.~Debnath, R.~Hu, D.~Suris, C.~Ryali, K.~V.
  Alwala, H.~Khedr, A.~Huang, J.~Lei, T.~Ma, B.~Guo, A.~Kalla, M.~Marks,
  J.~Greer, M.~Wang, P.~Sun, R.~Rädle, T.~Afouras, E.~Mavroudi, K.~Xu, T.-H.
  Wu, Y.~Zhou, L.~Momeni, R.~Hazra, S.~Ding, S.~Vaze, F.~Porcher, F.~Li, S.~Li,
  A.~Kamath, H.~K. Cheng, P.~Dollár, N.~Ravi, K.~Saenko, P.~Zhang, and
  C.~Feichtenhofer, ``Sam 3: Segment anything with concepts,'' 2025. [Online].
  Available: \url{https://arxiv.org/abs/2511.16719}
\BIBentrySTDinterwordspacing

\bibitem{Geiger12cvpr}
A.~Geiger, P.~Lenz, and R.~Urtasun, ``Are we ready for autonomous driving? the
  {KITTI} vision benchmark suite,'' in \emph{IEEE Conf. on Computer Vision and
  Pattern Recognition (CVPR)}, Providence, USA, June 2012, pp. 3354--3361.

\end{thebibliography}

\end{document}